\documentclass[11pt,letterpaper]{article}
\usepackage[margin=1in]{geometry}
\usepackage[T1]{fontenc}
\usepackage[utf8]{inputenc}
\usepackage{times}

\usepackage{amsmath,amsfonts,bm}

\def\eqref#1{equation~\ref{#1}}

\def\1{\bm{1}}

\DeclareMathAlphabet{\mathsfit}{\encodingdefault}{\sfdefault}{m}{sl}
\SetMathAlphabet{\mathsfit}{bold}{\encodingdefault}{\sfdefault}{bx}{n}

\usepackage{microtype}
\usepackage{graphicx}
\usepackage{wrapfig}
\usepackage[table]{xcolor}
\usepackage{array,multirow,makecell,tabularx,booktabs}
\usepackage{arydshln}
\usepackage{subcaption}
\usepackage{tcolorbox}
\usepackage{pifont}
\usepackage{enumitem}
\usepackage{stag_arxiv}
\usepackage{bookmark}
\definecolor{lightgray}{gray}{.92}
\makeatletter
\newcommand{\thickhline}{%
  \noalign{\ifnum0=`}\fi\hrule height 0.7pt
  \futurelet\reserved@a\@xhline
}
\makeatother
\title{\bfseries STaG: Human-Annotation-Free Unsupervised Domain Adaptation for Text-Based Person Retrieval}
\author{%
Yucheng Ji\textsuperscript{1} \quad
Yang Bai\textsuperscript{2} \quad
Yang Li\textsuperscript{1} \quad
Xinyu Zhou\textsuperscript{1} \quad
Min Cao\textsuperscript{1} \\[6pt]
\small \textsuperscript{1}School of Computer Science and Technology, Soochow University, Suzhou, China \\
\small \textsuperscript{2}School of Computer Science, Wuhan University, Wuhan, China
}

\date{}
\hypersetup{
  pdftitle={STaG: Human-Annotation-Free Unsupervised Domain Adaptation for Text-Based Person Retrieval},
  pdfauthor={Yucheng Ji, Yang Bai, Yang Li, Xinyu Zhou, Min Cao},
  pdfsubject={Text-based person retrieval and unsupervised domain adaptation},
  pdfkeywords={text-based person retrieval, unsupervised domain adaptation, STaG}
}
\begin{document}
\maketitle

\begin{abstract}
    Real-world deployment of Text-Based Person Retrieval (TBPR) is bottlenecked by the absence of textual annotations in new domains.
    Existing Unsupervised Domain Adaptation (UDA-TBPR) methods still assume access to target-domain human-written annotations (albeit unpaired with images), limiting their practicality.
    We consider a more practical setting: human-annotation-free UDA-TBPR, with labeled image–text data in the source domain and annotation-free, unlabeled images in the target domain.
    This setting introduces two core challenges: the absence of cross-modal alignment in the target domain, and visual domain shifts from source to target.
    To tackle these, we propose STaG, a dual-path framework comprising Style-Agnostic Text Alignment (SATA) and Heterogeneous Graph Memory Alignment (HGMA). Specifically, SATA leverages source-domain data to decouple identity-discriminative semantics from machine-generated stylistic artifacts, establishing robust cross-modal alignment for the target domain. Meanwhile, HGMA exploits cross-domain semantic neighborhoods via a memory-based graph to learn domain-invariant visual features. Ultimately, STaG achieves state-of-the-art performance across three benchmarks under this realistic setting.
\end{abstract}

\begin{wrapfigure}{r}{0.5\linewidth}
    \vspace{-16pt}
    \centering
    \includegraphics[
        width=\linewidth,
        trim=0 7 0 0,
        clip
    ]{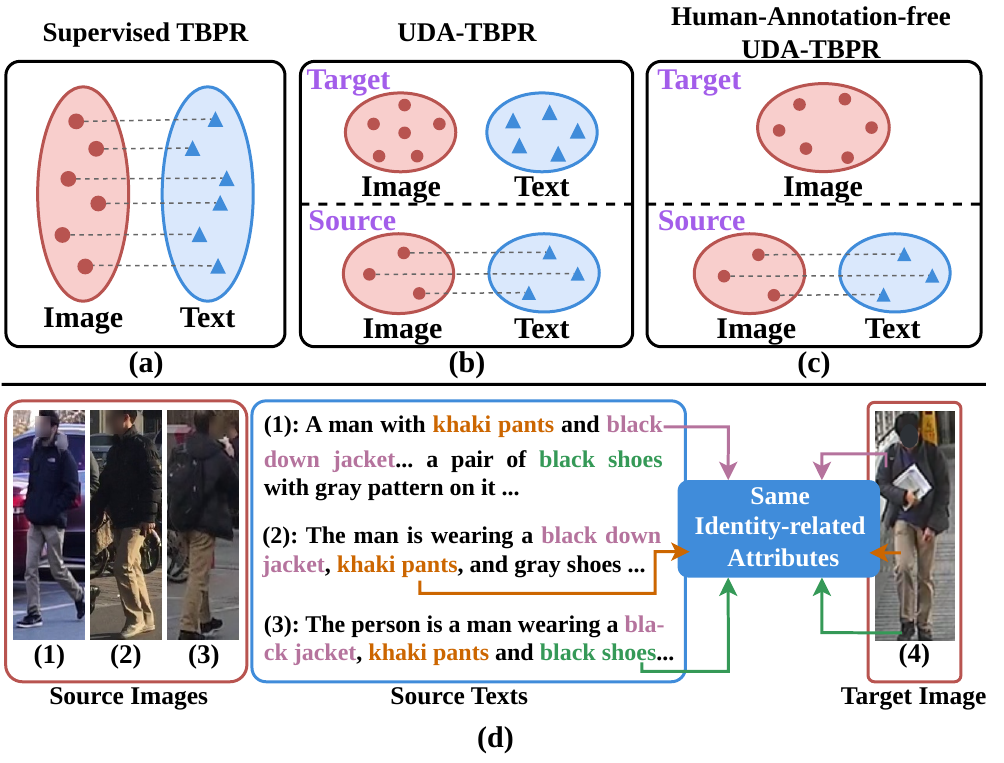}
    \caption{Comparison of different supervision paradigms in TBPR. Our setting relies strictly on unlabeled target images.}
    \label{fig:intro}
\end{wrapfigure}

\vspace{-8pt}
\section{Introduction}
\label{sec:intro}
\vspace{-4pt}

Text-Based Person Retrieval (TBPR)~\citep{li2025exploring,bai2023rasa,cvpr23crossmodal} aims to retrieve images of a specific individual from a gallery given a textual description.
Compared to general image–text retrieval~\citep{radford2021learning, li2021align, faghri2018vse}, which typically involves matching across diverse semantic categories, TBPR is uniquely challenging as it requires fine-grained discrimination among highly similar human identities.
This task is complicated by significant intra-class variation (\emph{e.g.}, changes in pose, clothing, or lighting for the same person) and limited inter-class distinction (\emph{e.g.}, visual similarity between different individuals).
Despite these challenges, TBPR has attracted growing interest~\citep{zhang2018deep, cvpr23crossmodal, qin2024noisy, tan2024harnessing} due to its high practical value in public safety scenarios, including assisted searches for suspects or missing persons in surveillance footage.

Most existing approaches to TBPR operate in a fully supervised setting, relying on large-scale datasets comprising collected images paired with carefully curated human annotations (Fig.~\ref{fig:intro} (a)).
However, in practical deployment scenarios, while image galleries are readily available from surveillance streams, acquiring aligned human-written text annotations for each identity is prohibitively expensive and labor-intensive.
This reliance on paired supervision severely hinders the scalability of current methods to new environments and poses a significant barrier to real-world adoption.
A promising direction is to introduce unsupervised domain adaptation into TBPR (UDA-TBPR), where cross-modal knowledge is transferred from a labeled source domain to the unlabeled target domain.
However, existing UDA-TBPR methods~\citep{jing2020cross, Luo2025GraphBasedCK, NIU2026112521} typically assume that human-written textual descriptions for the target domain are available, only unpaired with images (Fig.~\ref{fig:intro} (b)).
This assumption, however, is often unrealistic.
In real-world scenarios, person-specific textual descriptions do not exist independently; they are derivative of visual inspection, explicitly generated by annotators observing specific identities.
Therefore, the existence of target-domain human texts implies that a latent image-text correspondence has already been established manually.
Consequently, \textit{these methods still inevitably incur prohibitive manual annotation costs, failing to provide a truly human-annotation-free solution.}
This dependence on manual annotation remains a critical bottleneck that impedes practical deployment.

Motivated by this observation, we consider a more practical, human-annotation-free UDA-TBPR setting: the source domain provides paired image-text data, while the target domain contains only unlabeled images (as shown in Fig.~\ref{fig:intro} (c)).
This setting, while realistic, introduces two substantial challenges:
\textbf{\ding{182} Lack of Cross-modal Alignment.}
The absence of explicit textual anchors in the target domain leaves the joint embedding space severely under-constrained, making it exceptionally difficult to associate target visual identities with textual queries.
\textbf{\ding{183} Visual Domain Shift.}
Differences in imaging conditions, backgrounds, resolution, and camera styles make representations overfit to source-specific biases, leading to a catastrophic drop in retrieval accuracy when deployed in new environments.

To address these challenges, we propose STaG (Style-agnostic Text and Graph memory alignment), a novel framework that reconstructs cross-modal semantic associations and mitigates cross-domain distribution shifts by leveraging labeled source-domain data and unlabeled target-domain images.
Specifically, STaG integrates two complementary components.
\textbf{\ding{182} Style-Agnostic Text Alignment (SATA).}
To enable explicit cross-modal alignment in the target domain, a prevailing approach is to utilize MLLMs to synthesize pseudo-descriptions that compensate for the missing modality~\citep{tan2024harnessing, bai2023text}. However, relying exclusively on these generated texts often degrades retrieval robustness, as their stylistic divergence from human-written queries causes the model to overfit to rigid, machine-specific phrasing (as shown in Fig.~\ref{fig:app_sata}).
To mitigate this, SATA introduces a style-invariance learning paradigm that exploits the source domain to teach the model how to decouple underlying identity semantics from superficial textual styles. In this way, the model develops a style-agnostic representation space that enables the framework to utilize target-domain pseudo-texts without being corrupted by generation artifacts.
\textbf{\ding{183} Heterogeneous Graph Memory Alignment (HGMA).}
To mitigate visual domain shift, we emphasize domain-invariant person cues by leveraging a semantic similarity prior: visually similar images tend to share identity-related attributes (Fig.~\ref{fig:intro} (d)).
Guided by this prior, we can dynamically inject source-domain visual features into target-domain representations to bridge the domain gap. To enhance representational robustness, we introduce a graph structure to capture broader neighborhood topologies simultaneously.
Distinct from prior memory-based and graph-based methods~\citep{ahmadi2020memory, du2023joint, Luo2025GraphBasedCK} that operate within a single modality or require target-domain human-annotations,
HGMA constructs a heterogeneous graph that links each target image to its most relevant cross-domain neighbors, allowing the target representation to overcome domain-specific biases through aggregated neighbor support.
Furthermore, to shield this aggregation from noisy neighbors, we introduce a Confidence-Aware Entropy regularizer (CAE-Loss). Rather than minimizing entropy over single-modality classifier logits~\citep{yue2026beyond,yang2023trust, zhang2022confidence}, CAE sharpens cross-modal matching distributions to ensure high-quality image-text alignments.

Our contributions can be summarized as follows:

\ding{182} We introduce a practical, human-annotation-free UDA-TBPR setting. By entirely eliminating the reliance on target-domain human annotations, this flexible paradigm offers a more realistic solution for real world deployment.

\ding{183} We propose STaG, a novel framework tailored for our new challenging setting. It incorporates SATA to establish robust cross-modal associations and HGMA to mitigate visual domain shifts, achieving robust alignment without target-domain human annotations.

\ding{184} Extensive experiments across three standard benchmarks show that STaG outperforms state-of-the-art methods and effectively narrows the gap to fully supervised upper bounds, validating its robustness entirely without target-domain human annotations.

\vspace{-12pt}
\section{Related Work}
\vspace{-6pt}
\subsection{Supervised TBPR}
\vspace{-4pt}

TBPR aims to retrieve images of a target person from a gallery given a natural-language description. Since the task was introduced by \citep{li2017person} together with the CUHK-PEDES benchmark, TBPR has evolved into a representative problem at the intersection of person re-identification and vision-language retrieval.
Most supervised TBPR methods can be grouped into two lines.
\ding{182} Cross-modal alignment methods learn a shared embedding space for images and texts.
Early approaches primarily relied on global feature alignment~\citep{li2017person, zheng2020dual, farooq2020convolutional}, while subsequent work explored fine-grained correspondences across parts, phrases, or multiple granularities~\citep{chen2022tipcb, suo2022simple, wu2024laip, qi2025granularity, li2024adaptive}.
Many of these methods introduce structural priors, such as human parsing, pose estimation, and linguistic toolkits (e.g., NLTK~\citep{loper2002nltk}), to better associate local visual cues with textual entities~\citep{loper2002nltk, wang2020vitaa, jing2020pose, zheng2020hierarchical}.
\ding{183} Representation learning methods target modality-invariant identity features and robustness under nuisance factors.
Typical strategies include auxiliary objectives (e.g., attribute/color reasoning) and debiasing modules (e.g., reducing over-reliance on color) to improve cross-modal discriminability and generalization~\citep{wu2021lapscore, wang2022caibc}, often coupled with adaptive alignment mechanisms to handle complex scenes~\citep{li2022learning, gao2021contextual, gao2022conditional, ji2022asymmetric}.
Recent studies further leverage pretrained vision-language models (VLMs), such as CLIP~\citep{radford2021learning}, and multimodal LLMs via prompt learning, feature calibration, or knowledge transfer to enhance retrieval performance~\citep{cvpr23crossmodal, han2021text, yan2022clip, wei2023calibrating, cao2024empirical, lu2025prompt, bai2023rasa}. Most recently, the community has explored generative and interactive paradigms. For instance, DSEE~\citep{Toward2026chen} employs generative-driven semantic enhancement to enrich cross-modal semantic diversity. Other work incorporates interactive querying and contextual cues to refine query-image matching~\citep{qin2025human, bai2025chat, xu2025sa}. Despite these advances, most existing approaches (including VLM-based variants) still rely heavily on paired image--text supervision during training and/or adaptation. When the target domain provides only unlabeled images and no textual descriptions, target images cannot directly participate in cross-modal alignment, making such methods difficult to deploy under
a more practical human-annotation-free UDA-TBPR setting.

\vspace{-6pt}
\subsection{Unsupervised Domain Adaptation for TBPR}
\vspace{-4pt}

To mitigate cross-domain distribution shifts, several works have explored Unsupervised Domain Adaptation (UDA) for TBPR. \citep{niu2019fusing} proposed fusing direction-specific adaptation strategies for real-world person search; MAN~\citep{jing2020cross} addressed the domain gap by aligning feature distributions through moment matching; and GCKD~\citep{Luo2025GraphBasedCK} constructed cross-modal graphs to propagate relational structures and distill knowledge across domains.
However, this graph construction is similarly predicated on the availability of target-domain descriptions to establish the necessary nodes and edges for message passing.
Crucially, these methods assume access to target-domain human annotations during adaptation.
Such annotations are costly and often impractical to collect in real-world surveillance, motivating a strictly human-annotation-free paradigm.

\vspace{-6pt}
\subsection{Low Resource TBPR}
\vspace{-4pt}

Beyond UDA-TBPR, low-resource TBPR addresses scenarios where the target domain itself is data-scarce and lacks a designated source domain. These works mainly fall into two categories.
Weakly supervised and semi-supervised TBPR reduce identity-level supervision by assuming paired image--text data without identity labels, or a small number of labeled pairs together with abundant unlabeled data~\citep{zhao2021weakly, zheng2024cpcl, fu2025similarity, gong2024enhancing, zhang2025dual, gao2024semi}.
Despite reduced labeling requirements, these methods still fundamentally rely on paired cross-modal data and typically require textual descriptions to enable effective cross-modal alignment.
Fully unsupervised TBPR further
eliminates the need for both identity labels and explicit image–text pairings
and instead relies on pseudo texts generated or rewritten by VLMs/MLLMs and then learning with noise-robust objectives~\citep{bai2023text, Shao2023UnifiedPW, ijcai2024p116, mu2025face, ye2026gtr, niu2026pseudo, chen2026unsuper}. While reducing annotation cost, performance is often constrained by pseudo-text quality and noise in the induced supervision.
Different from these works, we explicitly leverage the rich, well-annotated resources of the source domain to enhance TBPR performance in the target domain, thereby bridging the gap between cross-domain knowledge transfer and low-resource retrieval.

\begin{figure}[t]
    \centering

    \includegraphics[width=0.92\linewidth]{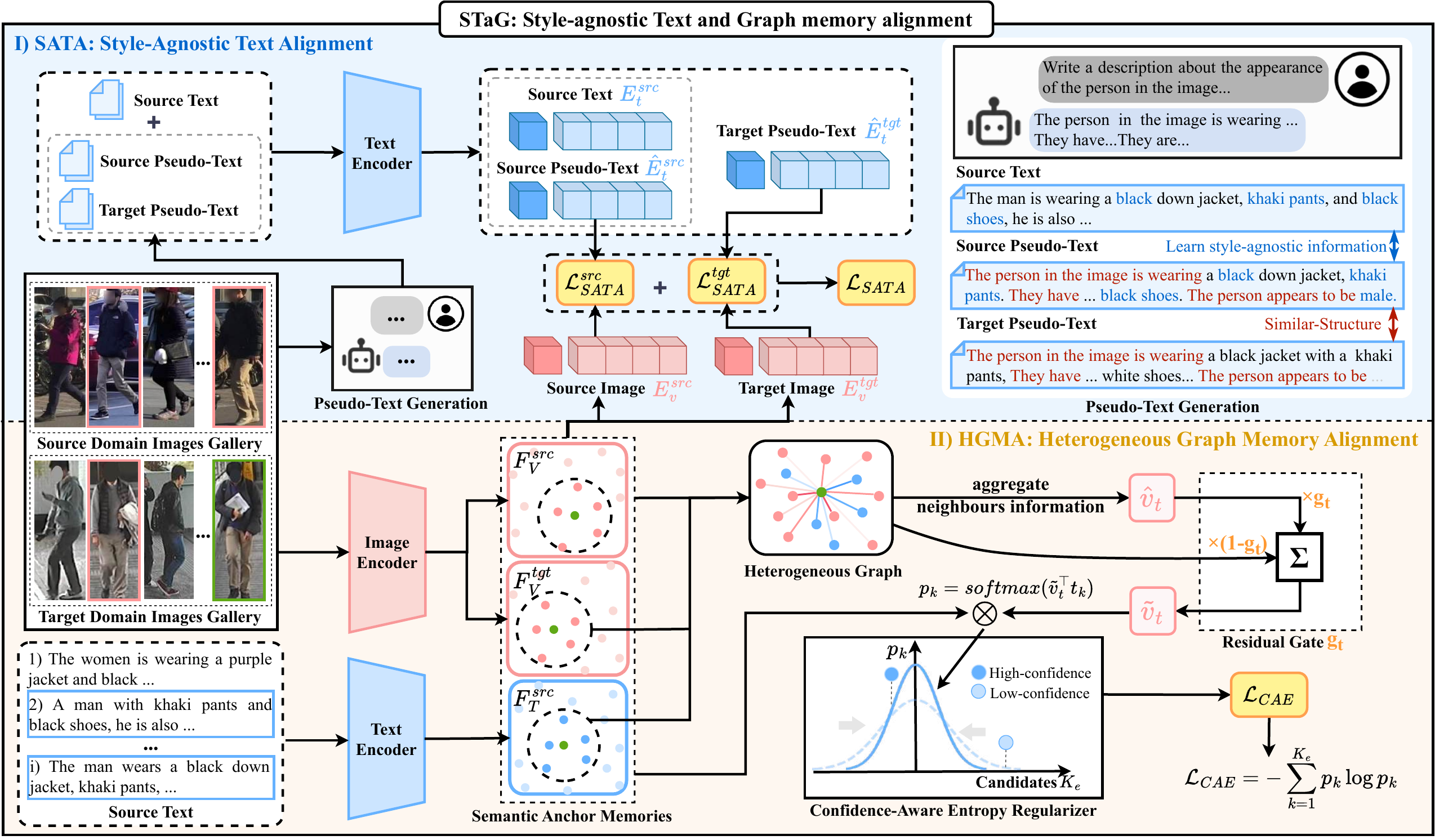}
     \caption{Overview of the proposed STaG. It consists of SATA and HGMA. \textbf{SATA} learns style-agnostic text alignment by enforcing identity-level consistency between human-written source descriptions and model-generated pseudo descriptions, producing reliable pseudo-text supervision for target-domain retrieval. \textbf{HGMA} mitigates visual domain shift by constructing a heterogeneous graph over cross-domain nearest neighbors and aggregating neighborhood information via semantic anchor memories, with residual gating $g_t$ and a confidence-aware entropy regularizer $\mathcal{L}_{CAE}$ to improve representation robustness.}
    \label{fig:model}
    \vspace{-12pt}
\end{figure}

\vspace{-12pt}
\section{Method}
\label{sec:Method}
\vspace{-6pt}

In this section, we present our framework STaG under a more practical UDA-TBPR setting, where the source domain contains paired image–text data and the target domain contains only unlabeled images. Let the source domain be denoted as $D^s=\{(I_i^s,T_i^s)\}_{i=1}^{N_s}$ and the target domain as $D^t=\{I_i^t\}_{i=1}^{N_t}$, where $I$ is a pedestrian image and $T$ is its textual description. As illustrated in Fig.~\ref{fig:model}, STaG comprises two complementary components: Style-Agnostic Text Alignment (SATA) and Heterogeneous Graph Memory Alignment (HGMA).

\vspace{-6pt}
\subsection{Preliminary}
\label{sec:pre}
\vspace{-4pt}

We adopt a CLIP-style dual-encoder as default retrieval backbone; it can also be replaced with other backbones as verified in Section~\ref{sec:compare-sota}.
The backbone comprises an image encoder $f_\theta(\cdot)$ and a text encoder $g_\phi(\cdot)$. Given a mini-batch of $M$ paired samples $\{(I_i,T_i)\}_{i=1}^{M}$, we compute $\ell_2$-normalized embeddings as follows:
\begin{equation}
v_i = \frac{f_\theta(I_i)}{\|f_\theta(I_i)\|_2},\qquad
t_i = \frac{g_\phi(T_i)}{\|g_\phi(T_i)\|_2}.
\label{eq:def_vt}
\end{equation}
A symmetric contrastive loss (InfoNCE) is employed to pull matched pairs together and push unmatched pairs away:
\begin{equation}
\mathcal{L}_{i\rightarrow t}
= -\frac{1}{M}\sum_{i=1}^{M}\log
\frac{\exp(v_i^\top t_i / \tau)}{\sum_{j=1}^{M}\exp(v_i^\top t_j / \tau)},
\quad
\mathcal{L}_{t\rightarrow i}
= -\frac{1}{M}\sum_{i=1}^{M}\log
\frac{\exp(t_i^\top v_i / \tau)}{\sum_{j=1}^{M}\exp(t_i^\top v_j / \tau)},
\label{eq:clip-t2i}
\end{equation}
where $\tau$ is a temperature coefficient.
The overall contrastive training objective is formulated as:
\begin{equation}
\mathcal{L}_{itc}=\frac{1}{2}\left(\mathcal{L}_{i\rightarrow t}+\mathcal{L}_{t\rightarrow i}\right).
\label{eq:clip-itc}
\end{equation}

\vspace{-6pt}
\subsection{SATA: Style-Agnostic Text Alignment}
\label{sec:sata}
\vspace{-4pt}

In the absence of target-domain textual descriptions, target images cannot participate in explicit image–text alignment signals.
While generating pseudo-descriptions for target images offers a potential remedy, machine-generated texts often exhibit a severe stylistic divergence compared to human-written queries (as shown in Fig.~\ref{fig:app_sata}(b)). This style mismatch can introduce bias and ultimately degrade retrieval robustness.

To address this challenge, SATA leverages source-domain data to establish a style-invariance constraint. The core objective is to enforce representational consistency between human-written queries and machine-generated descriptions.
By aligning both the original human annotation and a generated pseudo-description to the identical source visual anchor, the model is inherently penalized for relying on surface-level phrasing (Fig.~\ref{fig:app_sata}(b)). Instead, it is compelled to distill style-invariant, identity-discriminative semantic features.
Crucially, since the pseudo-texts in both domains are synthesized using the same generator $G(\cdot)$ and prompt template (Fig.~\ref{fig:app_sata}(a)), these machine-generated descriptions remain strictly consistent across domains. Consequently, the style invariance calibrated on the source domain seamlessly transfers to the target domain, enabling robust cross-modal alignment even in the absence of human annotations.
Specifically, using a unified prompt template (detailed in Appendix~\ref{app:prompt}), we synthesize pseudo-descriptions for both domains as $\hat{T}_{i}^{s}=G(I_{i}^{s})$ and $\hat{T}_{j}^{t}=G(I_{j}^{t})$.
We then construct augmented datasets to facilitate joint optimization.
For the source domain, we pair each image with both its original human-written text \(T^s\) and its generated pseudo-text \(\hat{T}^s\). For the target domain, we utilize the generated pairs, yielding the following sets for contrastive training:
\begin{equation}
\hat{\mathcal{D}}^s=\mathcal{D}^s\bigcup\left\{\langle I_i^s, \hat{T}_i^s\rangle\right\}_{i=1}^{N_s},\quad
\hat{\mathcal{D}}^t=\left\{\langle I_i^t,   \hat{T}_i^t\rangle\right\}_{i=1}^{N_t},
\end{equation}
where $\hat{\mathcal{D}}^s$ denotes the augmented source dataset which combines the ground-truth pairs in $\mathcal{D}^s$ with generated image-pseudo-text pairs $\langle I, \hat{T}\rangle$ and $\hat{\mathcal{D}}^t$ represents the target dataset constructed entirely from such generated pairs.
The retrieval backbone is then optimized using the SATA objective:
\begin{equation}
\mathcal{L}_{\mathrm{SATA}}=\mathcal{L}^{src}_{\mathrm{SATA}}+\mathcal{L}^{tgt}_{\mathrm{SATA}}
=\frac{1}{|\hat{\mathcal{D}}^s|}\sum_{(I,T)\in\hat{\mathcal{D}}^s}\mathcal{L}_{itc}(I,T)+\frac{1}{|\hat{\mathcal{D}}^t|}\sum_{(I,T)\in\hat{\mathcal{D}}^t}\mathcal{L}_{itc}(I,T),
\end{equation}
where $\mathcal{L}_{itc}$ is the symmetric image-text contrastive loss defined in Eq.~\ref{eq:clip-itc}.
By minimizing \(\mathcal{L}^{src}_{\mathrm{SATA}}\), the shared visual anchor acts as a critical semantic filter, decoupling style-invariant semantic representations from rigid sentence structures. Crucially, since the target-domain pseudo-texts contain these exact same generation artifacts, optimizing $\mathcal{L}_{SATA}^{tgt}$ allows the pre-calibrated text encoder to intrinsically filter out the machine-style biases. Rather than overfitting to the generated phrasing, $\mathcal{L}_{SATA}^{tgt}$ effectively projects the target images into the unified, style-agnostic semantic space, achieving robust adaptation despite the complete lack of human annotations.
To empirically substantiate this decoupling mechanism, we provide extensive visual analyses in Appendices~\ref{app: distribution} and \ref{app:att in text}.

Notably, since the target domain strictly lacks human annotations, synthesizing pseudo-descriptions via MLLM serves as a critical enabler for explicit cross-modal alignment. Crucially, this generation is strictly a one-time, offline preprocessing step decoupled from the training loop, introducing zero additional latency. A detailed cost analysis in Appendix~\ref{app:mllm costs} demonstrates that STaG is a practical and scalable solution highly suitable for real-world deployment.

\vspace{-6pt}
\subsection{HGMA: Heterogeneous Graph Memory Alignment}
\label{method: hgma}
\vspace{-4pt}

Due to inherent visual domain shifts, an encoder trained only on the source domain inevitably produces shifted embeddings for target images, degrading cross-domain matching performance.
To mitigate this, we propose the Heterogeneous Graph Memory Alignment (HGMA) module.
HGMA leverages the insight that visually similar samples share identity-related semantics regardless of their domain. Consequently, it constructs a heterogeneous graph to aggregate semantic cues from both cross-domain and intra-domain neighbors, yielding a more robust target representation.
Distinct from prior memory- and graph-based methods~\citep{ahmadi2020memory, du2023joint, Luo2025GraphBasedCK} that either propagate evidence within a single modality or require target-domain human-annotations, both of which are structurally inapplicable under our human-annotation-free setting.

Given a target image feature $v_t$, relying solely on candidate anchors within a single mini-batch is insufficient.
To expand the semantic search space, we maintain three fixed-size First-In-First-Out (FIFO) Semantic Anchor Memories (SAMs): $\{\mathcal{M}_v^s, \mathcal{M}_t^s, \mathcal{M}_v^t\}$, which cache the encoded features of source images, source texts, and target images, respectively.
These memories are updated dynamically during training via an enqueue-dequeue mechanism
\footnote{Bounded by a constant capacity $M$, the space complexity of each SAM is strictly $\mathcal{O}(M \times D)$ (where $D$ is the feature dimension).
This decouples the memory overhead from the overall dataset scale, keeping the memory cost bounded during training. For a detailed complexity analysis, see Appendix~\ref{app: train complex}.}.
Leveraging these SAMs, we retrieve the most relevant semantic anchors to populate the graph nodes. Specifically, the intra-domain neighbor set $\mathcal{N}_v^t$ is formulated as the optimal subset of size $K_g$ that maximizes the cosine similarity score:
\begin{equation}
    \mathcal{N}_v^t = \underset{\mathcal{S} \subset \mathcal{M}_v^t, |\mathcal{S}|=K_g}{\arg\max} \sum_{m \in \mathcal{S}} \frac{v_t^\top m}{\|v_t\| \|m\|}.
    \label{eq:topk_retrieval}
\end{equation}
Ideally, $\mathcal{N}_v^t$ captures the local manifold structure in the target domain. Similarly, we independently obtain cross-domain neighbor sets $\mathcal{N}_v^s$ and $\mathcal{N}_t^s$ from $\mathcal{M}_v^s$ and $\mathcal{M}_t^s$, respectively.
The final neighborhood $\mathcal{N}=\{\mathcal{N}_v^s, \mathcal{N}_t^s, \mathcal{N}_v^t\}$ comprises a total of $3 \times K_g$ anchors, serving as the basis for graph aggregation.
Following neighbor retrieval, we compute an attention-based aggregation for each set.
For any $\mathcal{N}_x \in \{\mathcal{N}_v^s, \mathcal{N}_t^s, \mathcal{N}_v^t\}$, we~define:
\begin{equation}
\mathrm{Agg}(v_t;\mathcal{N}_x)
=\sum_{n\in\mathcal{N}_x}
\underbrace{
\frac{\exp(v_t^\top n/\tau_g)}{\sum_{n'\in\mathcal{N}_x}\exp(v_t^\top n'/\tau_g)}}_{w_{t,n}^{(x)}}\cdot n,
\label{eq:agg_def}
\end{equation}
where $w_{t,n}^{(x)}$ denotes the normalized attention weight assigned to neighbor $n$.
We then fuse $v_t$ with the aggregated messages via set-specific linear projections:
\begin{equation}
    \hat{v}_t
    =\sigma\big(
    W_0 v_t +\sum_{\mathcal{N}_x}W_{\mathcal{N}_x}\,\mathrm{Agg}\!\left(v_t;\mathcal{N}_x\right)\big),
    \label{eq:hetero_fuse}
\end{equation}
where $\sigma(\cdot)$ is the $\mathrm{ReLU}$ activation function, $W_0$ and $W_{\mathcal{N}_x}$ are learnable linear projections. The use of separate projections $W_{\mathcal{N}_x}$ allows the model to account for heterogeneous statistics across the different memory sources while fusing them into a unified embedding space.

To prevent negative transfer from weakly related neighbors, we apply a confidence-based residual gate $g_t\in[0,1]$. We define confidence as the mean similarity between $\hat{v}_t$ and all retrieved neighbors, mapped to $[0,1]$ via a sigmoid function $\phi(\cdot)$:
\begin{equation}
g_t=\phi\left(\frac{1}{|\mathcal{N}|}\sum_{n\in\mathcal{N}}\hat{v}_t^\top n\right), \quad
\tilde{v}_t=\mathrm{norm}\left((1-g_t)v_t+g_t\hat{v}_t\right),
\end{equation}
where $\mathrm{norm}(\cdot)$ denotes $\ell_2$ normalization, resulting in the corrected target embedding $\tilde{v}_t$.
We further introduce a Confidence-Aware Entropy (CAE) regularizer that sharpens a cross-modal matching distribution over retrieved source-text anchors,
distinct from classifier-entropy minimization in test-time adaptation, which operates on closed-set single-modality logits\citep{yue2026beyond,yang2023trust, zhang2022confidence}.
Specifically, given $\tilde{v}_t$, we retrieve its Top-$K_e$ nearest source-text embeddings $\{t_k\}_{k=1}^{K_e}$ from $\mathcal{M}_t^s$, compute a matching distribution $p_k$, and minimize its entropy:
\begin{equation}
p_k=\frac{\exp(\tilde{v}_t^\top t_k / \tau_e)}{\sum_{j=1}^{K_e}\exp(\tilde{v}_t^\top t_j / \tau_e)},
\quad
\mathcal{L}_{\mathrm{CAE}}=-\sum_{k=1}^{K_e} p_k\log p_k .
\label{eq:cae}
\end{equation}
where $\tau_e$ is a temperature parameter.
Minimizing $\mathcal{L}_{\mathrm{CAE}}$ sharpens the matching distribution, thereby encouraging high-confidence image-text associations and reducing ambiguous alignments.
Our overall training objective is then given by:
\begin{equation}
\mathcal{L}=\mathcal{L}_{\mathrm{SATA}}+\mathcal{L}_{\mathrm{CAE}}.
\label{eq:loss}
\end{equation}

\vspace{-12pt}
\section{Experiments}
\vspace{-6pt}

In this section, we evaluate the effectiveness of the proposed STaG framework. Comprehensive details regarding the datasets, protocols, and implementation details are provided in Appendix~\ref{app: exp setup}.

\vspace{-6pt}
\subsection{Comparisons with State-of-The-Art Methods}
\label{sec:compare-sota}
\vspace{-4pt}

To evaluate the effectiveness of the proposed STaG framework, we conduct the comparison with state-of-the-art methods under the human-annotation-free UDA-TBPR setting.
Notably, existing methods are primarily designed for either fully supervised or human-annotation-available UDA-TBPR settings.
To ensure a fair and meaningful comparison under our human-annotation-free UDA-TBPR protocol, we reproduced representative existing methods using the identical MLLM-generated pseudo-texts $\hat{T}^t$.
Quantitative results are summarized in Table~\ref{tab:merged_comparison}.
Key observations are outlined below:
\ding{182} STaG significantly outperforms all compared methods. Notably, when instantiated with RaSa as the baseline, STaG establishes a new performance upper bound, achieving 50.45\% R-1 accuracy on the challenging CUHK-PEDES $\rightarrow$ ICFG-PEDES.
The inferior performance of the compared methods can be attributed to their design assumptions. Originally developed for fully supervised setting or annotation-available UDA-TBPR, these methods lack mechanisms specifically tailored to the more practical and challenging human-annotation-free scenario. Consequently, they struggle to achieve sufficient cross-modal alignment in the absence of target-domain annotations.
On the other hand, fully supervised models (IVT, CFine, RICA, IRRA, TBPS-CLIP, and RaSa) are unable to handle the visual domain shift inherent in UDA-TBPR.
In contrast, STaG explicitly addresses both challenges through its proposed SATA and HGMA modules.
\ding{183} STaG achieves consistent performance gains across diverse baselines, including IRRA, TBPS-CLIP, and RaSa, showing its broad applicability and strong compatibility. For instance, when applied to the IRRA baseline in ICFG-PEDES $\rightarrow$ CUHK-PEDES, STaG improves R@1 by 11.21\%, underscoring its effectiveness across varied architectural designs.
\ding{184} Furthermore, extended evaluations in Appendix~\ref{app:appendix_experiments} demonstrate that despite operating under the most stringent human-annotation-free setting, STaG secures state-of-the-art performance across all datasets. Notably, it not only effectively bridges the gap to fully supervised upper bounds but also surpasses traditional text-dependent UDA methods in specific cross-domain scenarios.

\begin{table}[t]
\centering
\caption{
\textbf{Comparison with state-of-the-art methods under the human-annotation-free UDA-TBPR setting.}
We evaluate performance across four transfer scenarios, using CUHK-PEDES and ICFG-PEDES as source domains, respectively.
The table compares other methods (top) with our \textbf{STaG} module integrated into three backbones (bottom).
Absolute performance Improvements ($\Delta$) are highlighted in \textbf{bold}.
Note: all compared methods were originally designed for data-rich settings that assume access to textual annotations. For fair comparison, we reproduced them under our more practical, human-annotation-free UDA-TBPR setting.
}
\label{tab:merged_comparison}
\small
\setlength\tabcolsep{3pt}
\renewcommand\arraystretch{1.15}
\newcommand{\db}[1]{\textbf{+#1}}
\resizebox{0.95\textwidth}{!}{
\begin{tabular}{l cccc cccc c cccc cccc}
\hline\thickhline
\multirow{3.5}{*}{Method} & \multicolumn{8}{c}{\textbf{Source: CUHK-PEDES}} && \multicolumn{8}{c}{\textbf{Source: ICFG-PEDES}} \\
\cmidrule(lr){2-9} \cmidrule(lr){11-18}
 & \multicolumn{4}{c}{$\rightarrow$ ICFG-PEDES} & \multicolumn{4}{c}{$\rightarrow$ RSTPReid} && \multicolumn{4}{c}{$\rightarrow$ CUHK-PEDES} & \multicolumn{4}{c}{$\rightarrow$ RSTPReid} \\
\cmidrule(lr){2-5} \cmidrule(lr){6-9} \cmidrule(lr){11-14} \cmidrule(lr){15-18}
 & R@1 & R@5 & R@10 & mAP & R@1 & R@5 & R@10 & mAP && R@1 & R@5 & R@10 & mAP & R@1 & R@5 & R@10 & mAP \\
\hline
IVT~\citep{shu2023see}
& 27.49 & 47.80 & 57.09 & 16.03 & 39.35 & 67.60 & 76.85 & 30.70
&& 22.63 & 42.29 & 52.36 & 19.85 & 43.70 & 65.10 & 75.55 & 37.65 \\
POUF~\citep{Tanwisuth2023POUFPU}
& 32.31 & 54.83 & 64.13 & 16.90 & 39.40 & 63.25 & 70.50 & 15.89
&& 18.21 & 37.96 & 57.88 & 15.98 & 34.45 & 57.90 & 68.10 & 25.25 \\
ReCLIP~\citep{Hu2023ReCLIPRC}
& 37.39 & 55.90 & 63.88 & 18.64 & 44.40 & 66.80 & 76.95 & 31.36
&& 25.20 & 46.25 & 55.96 & 23.39 & 40.80 & 62.35 & 71.90 & 32.08 \\
CFine~\citep{yan2022clip}
& 38.81 & 57.27 & 65.46 & -     & 48.30 & 71.90 & 82.45 & -
&& 32.67 & 54.03 & 63.68 & 25.47 & 47.40 & 70.60 & 79.35 & 42.30 \\
RICA~\citep{bai2026rica}
& 40.01 & 57.87 & 64.52 & 15.89 & 47.80 & 70.65 & 79.65 & 34.52
&& 34.11 & 55.99 & 65.74 & 30.44 & 39.95 & 63.75 & 73.30 & 31.63\\
GCKD~\citep{Luo2025GraphBasedCK}
& 42.15 & 60.36 & 68.81 & 24.01 & 54.00 & 77.35 & 85.05 & 36.62
&& 38.33 & 59.51 & 68.30 & 35.56 & 49.15 & 72.35 & 79.95 & 38.81 \\
\hline

IRRA~\citep{cvpr23crossmodal}
& 36.00 & 58.69 & 68.88 & 19.34 & 46.90 & 70.40 & 79.60 & 35.42
&& 33.43 & 56.11 & 66.23 & 31.38 & 45.10 & 69.20 & 78.75 & 36.76 \\
\textbf{+STaG (Ours) }
& 42.59 & 61.20 & 69.15 & 21.38 & 52.85 & 78.20 & 85.25 & 39.91
&& 44.64 & 65.76 & 74.03 & 38.98 & 48.60 & 71.80 & 80.35 & 37.43 \\
\rowcolor[HTML]{D7F6FF}
\textbf{Improvement $\Delta$}
& \db{6.59} & \db{2.51} & \db{0.27} & \db{2.04} & \db{5.95} & \db{7.80} & \db{5.65} & \db{4.49}
&& \db{11.21} & \db{9.65} & \db{7.80} & \db{7.60} & \db{3.50} & \db{2.60} & \db{1.60} & \db{0.67} \\
\cdashline{1-18}[1pt/2pt]

TBPS-CLIP~\citep{cao2024empirical}
& 37.72 & 56.41 & 64.27 & 19.39 & 48.75 & 71.35 & 81.60 & 35.85
&& 35.99 & 56.94 & 66.23 & 32.07 & 44.50 & 67.00 & 76.85 & 33.45 \\
\textbf{+STaG (Ours) }
& 44.79 & 61.85 & 69.11 & 24.21 & 49.50 & 72.35 & 82.20 & 36.13
&& 44.23 & 65.09 & 73.08 & 38.64 & 53.70 & 75.00 & 82.60 & 40.15 \\
\rowcolor[HTML]{D7F6FF}
\textbf{Improvement $\Delta$}
& \db{7.07} & \db{5.44} & \db{4.84} & \db{4.82} & \db{0.75} & \db{1.00} & \db{0.60} & \db{0.28}
&& \db{8.24} & \db{8.15} & \db{6.85} & \db{6.57} & \db{9.20} & \db{8.00} & \db{5.75} & \db{6.70} \\
\cdashline{1-18}[1pt/2pt]

RaSa~\citep{bai2023rasa}
& 48.14 & 65.29 & 71.83 & 22.65 & 58.05 & 79.50 & 85.60 & 43.50
&& 48.65 & 69.90 & 76.53 & 42.03 & 55.00 & 73.65 & 81.55 & 46.18 \\
\textbf{+STaG (Ours)}
& 50.45 & 67.24 & 73.55 & 24.23 & 60.75 & 82.25 & 87.15 & 45.88
&& 57.44 & 76.13 & 83.25 & 50.97 & 55.70 & 76.95 & 84.15 & 46.88 \\
\rowcolor[HTML]{D7F6FF}
\textbf{Improvement $\Delta$}
& \db{2.31} & \db{1.95} & \db{1.72} & \db{1.58} & \db{2.70} & \db{2.75} & \db{1.55} & \db{2.38}
&& \db{8.79} & \db{6.23} & \db{6.72} & \db{8.94} & \db{0.70} & \db{3.30} & \db{2.60} & \db{0.70} \\

\hline\thickhline
\end{tabular}
}
\vspace{-12pt}
\end{table}

\vspace{-6pt}
\subsection{Ablation Study}
\label{sec:ablation}
\vspace{-4pt}

\subsubsection{Effectiveness of the STaG Framework}

STaG comprises two components: SATA and HGMA. To assess their individual contributions, we conduct an ablation study, summarized in Table~\ref{tab:main_ablation}. Our observations are as follows:
\ding{182} Activating either module alone consistently outperforms the baseline, demonstrating that both cross-modal alignment (via SATA) and domain-invariant visual learning (via HGMA) are beneficial in the absence of target-domain text annotations;
\ding{183} Combining both modules yields the best performance, confirming that they address complementary challenges. Specifically, SATA reconstructs missing semantic associations between modalities, while HGMA mitigates visual distribution shifts across domains. Their synergistic integration ultimately enables robust cross-domain adaptation under human-annotation-free condition.

\begin{table}[t]
\centering
\caption{\textbf{Ablation study of STaG components.}
\textbf{Bold} indicates the best performance, and the final configuration (Ours) is highlighted in blue.}
\label{tab:main_ablation}
\small
\setlength\tabcolsep{4pt}
\renewcommand\arraystretch{1.15}
\resizebox{0.75\textwidth}{!}{
\begin{tabular}{l cc cccc cccc}
\hline\thickhline
\multirow{2}{*}{Methods} & \multirow{2}{*}{HGMA} & \multirow{2}{*}{SATA}
 & \multicolumn{4}{c}{ICFG-PEDES$\to$CUHK-PEDES} & \multicolumn{4}{c}{ICFG-PEDES$\to$RSTPReid} \\
\cmidrule(lr){4-7} \cmidrule(lr){8-11}
 & & & R@1 & R@5 & R@10 & mAP & R@1 & R@5 & R@10 & mAP \\
\midrule
Baseline       &              &              & 35.99 & 56.94 & 66.23 & 32.07 & 44.50 & 67.00 & 76.85 & 33.45 \\
+ HGMA         & $\checkmark$ &              & 36.63 & 58.08 & 67.46 & 32.94 & 47.25 & 71.10 & 79.85 & 38.37 \\
+ SATA         &              & $\checkmark$ & 41.03 & 59.71 & 67.46 & 36.11 & 50.15 & 72.10 & 80.90 & 37.32 \\
\rowcolor[HTML]{D7F6FF}
\textbf{+ STaG (Ours)} & $\checkmark$ & $\checkmark$
 & \textbf{44.23} & \textbf{65.09} & \textbf{73.08} & \textbf{38.64}
 & \textbf{53.70} & \textbf{75.00} & \textbf{82.60} & \textbf{40.15} \\
\hline\thickhline
\end{tabular}}
\vspace{-6pt}
\end{table}

\begin{table}[t]
\centering
\caption{\textbf{Ablation of key components in SATA.}
Experiments use ICFG-PEDES as the source domain. $\hat{T}^n$, $\hat{T}^s$, and $\hat{T}^t$ denote the use of random-token noise, source pseudo-texts, and target pseudo-texts, respectively. \textbf{Bold} indicates the best performance, and our final configuration is highlighted in blue.}
\label{tab:ablation_sata}
\small
\setlength\tabcolsep{4pt}
\renewcommand\arraystretch{1.15}
\resizebox{0.70\textwidth}{!}{
\begin{tabular}{l ccc cccc cccc}
\hline\thickhline
\multirow{2}{*}{Methods} & \multirow{2}{*}{$\hat{T}^n$} & \multirow{2}{*}{$\hat{T}^s$} & \multirow{2}{*}{$\hat{T}^t$}
 & \multicolumn{4}{c}{ICFG-PEDES$\to$CUHK-PEDES} & \multicolumn{4}{c}{ICFG-PEDES$\to$RSTPReid} \\
\cmidrule(lr){5-8} \cmidrule(lr){9-12}
 & & & & R@1 & R@5 & R@10 & mAP & R@1 & R@5 & R@10 & mAP \\
\midrule
Baseline  &              &              &              & 36.63 & 58.08 & 67.46 & 32.94 & 47.25 & 71.10 & 79.85 & 38.37 \\
$M_0$     & $\checkmark$ &              &              & 38.50 & 60.00 & 68.47 & 33.24 & 48.30 & 72.20 & 79.95 & 38.40 \\
$M_1$     &              & $\checkmark$ &              & 39.27 & 61.49 & 70.52 & 35.49 & 50.70 & 74.00 & 81.30 & 38.92 \\
$M_2$     &              &              & $\checkmark$ & 40.39 & 62.23 & 71.09 & 36.22 & 51.30 & 74.55 & 82.00 & 39.11 \\
\rowcolor[HTML]{D7F6FF}
\textbf{$M_3$ (Ours)} & & $\checkmark$ & $\checkmark$
 & \textbf{44.23} & \textbf{65.09} & \textbf{73.08} & \textbf{38.64}
 & \textbf{53.70} & \textbf{75.00} & \textbf{82.60} & \textbf{40.15} \\
\hline\thickhline
\end{tabular}}
\vspace{-6pt}
\end{table}

\vspace{-6pt}
\subsubsection{Effectiveness of the SATA Component}
\label{ablation:sata}
\vspace{-4pt}

We further analyze the SATA component to validate its core design motivation: enabling explicit cross-modal alignment in the target domain while mitigating the stylistic discrepancy between human-annotated and machine-generated descriptions. To this end, Table~\ref{tab:ablation_sata} compares three ablated variants and the full model that employ different forms of textual descriptions for cross-modal alignment. We make the following key observations:
\ding{182} $M_0$ replaces pseudo-texts with unstructured random-token noise ($\hat{T}^n$). This confirms that SATA's efficacy strictly relies on valid visual-semantic grounding, rather than acting as a generic text-based regularizer.
\ding{183} $M_1$ shows that using source-domain generated text alone for explicit cross-modal alignment yields only negligible performance gains. This confirms that the stylistic refinement mechanism is effective only when applied to alignment signals derived from the target domain.
\ding{184} Using target pseudo-texts $\hat{T}^t$ alone ($M_2$) enables explicit cross-modal alignment and improves performance over the baseline. However, the gains remain limited. As highlighted in the introduction, directly relying on model-generated descriptions risks overfitting to surface-level phrasing, leaving the model vulnerable to the distributional mismatch between synthetic pseudo-texts and real human-annotated descriptions.
\ding{185} The full model ($M_3$) embodies our complete solution. Compared to $M_2$, it additionally incorporates source pseudo-texts ($\hat{T}^s$), leveraging source images as stylistic references. By enforcing consistency between human-written and model-generated descriptions in the source domain, the model is encouraged to ignore stylistic variations and instead focus on identity-discriminative semantic content. This effectively \textit{calibrates} the encoder, producing pseudo-text supervision that is robust to stylistic shifts and more reliable for cross-modal alignment in the target domain.
\ding{186} Distribution-level analysis and token-level visualizations in Appendices~\ref{app: distribution} and~\ref{app:att in text} further show that SATA successfully mitigates the model's overfitting to specific sentence structures.
\ding{187} We additionally examine pseudo-text quality, generator choice, attribute hallucinations, and target-domain test-image corruptions in Appendix~\ref{sec:noise_robustness}, where STaG remains robust across the evaluated textual and visual perturbations.
Empirical evidence consistently demonstrates that STaG exhibits remarkable resilience against these generation errors. This exceptional stability is inherently driven by our synergistic architecture: SATA anchors the shared semantic space using clean source annotations, while HGMA effectively dilutes individual target-domain textual errors through structural neighborhood consensus.

\vspace{-6pt}
\subsubsection{Effectiveness of the HGMA Component}
\vspace{-4pt}

We ablate the internal components of HGMA in Table~\ref{tab:ablation_hgma} to evaluate the contributions of heterogeneous memory cues and confidence-aware regularization. Our key observations are as follows:
\ding{182} $M_0$ replaces KNN retrieval with uniformly random neighbors($\mathcal{M}_{rad}$). This confirms HGMA's benefit stems from identity-consistent graph topology rather than generic feature smoothing.
\ding{183} $M_1$ and $M_3$ ablate the cross-domain visual anchors ($\mathcal{M}_v^s$) and intra-domain visual anchors ($\mathcal{M}_v^t$), respectively. Omitting either consistently degrades retrieval performance, indicating that source visual anchors bridge the visual domain shift while target visual anchors capture the local manifold structure of the unlabeled target distribution.
\ding{184} $M_2$ ablates the cross-domain textual anchors ($\mathcal{M}_t^s$), causing a notable drop on both transfers. This directly substantiates HGMA's central design, that source-text anchors serve as cross-modal references in the absence of target-domain captions. Without them, the heterogeneous graph loses its only direct text-side supervision and the cross-modal alignment signal collapses back to the unimodal regime.
Retrieved neighbors from the SAMs are visualized in Appendix~\ref{app:hgma_neighbors}.
\ding{185} $M_4$ removes confidence-based residual gate ($g_t$), and $M_5$ removes confidence-aware entropy regularizer ($\mathcal{L}_{CAE}$) that enhances cross-modal matching reliability. The subsequent performance drops emphasize their necessity as structural safeguards. By modulating feature fusion and sharpening the matching distribution, these mechanisms are crucial for suppressing negative transfer from weakly related neighbors and mitigating error propagation when relying on noisy target-domain pseudo-texts.
Appendix~\ref{app:cae_generalization} further evaluates CAE under direct transfer, with no observed degradation.
\ding{186} Ultimately, integrating all components achieves the highest accuracy, confirming that robust adaptation requires the synergistic combination of accurate neighborhood retrieval, heterogeneous memory cues, and confidence-aware regularization.

\begin{table}[t]
\centering
\caption{\textbf{Ablation of HGMA.} Evaluation of memory configurations and regularization mechanisms across two cross-dataset settings. Bold indicates the best performance, and our final configuration is highlighted in blue.}
\label{tab:ablation_hgma}
\small
\setlength\tabcolsep{6pt}
\renewcommand\arraystretch{1.15}
\resizebox{0.7\textwidth}{!}{
\begin{tabular}{l cccc cccc}
\hline\thickhline
\multirow{2}{*}{Methods} & \multicolumn{4}{c}{ICFG-PEDES$\to$CUHK-PEDES} & \multicolumn{4}{c}{ICFG-PEDES$\to$RSTPReid} \\
\cmidrule(lr){2-5} \cmidrule(lr){6-9}
 & R@1 & R@5 & R@10 & mAP & R@1 & R@5 & R@10 & mAP \\
\midrule
Baseline
& 41.03 & 59.71 & 67.46 & 36.11
& 50.15 & 72.10 & 80.90 & 37.32 \\
$M_0$ (STaG w/ $\mathcal{M}_{rad}$)
& 41.57 & 60.29 & 67.88 & 36.40
& 50.65 & 72.50 & 81.00 & 37.56 \\
$M_1$ (STaG w/o $\mathcal{M}_v^s$)
& 43.72 & 64.89 & 72.85 & 38.29
& 51.55 & 73.45 & 81.70 & 38.83 \\
$M_2$ (STaG w/o $\mathcal{M}_t^s$ )
& 42.88 & 63.90 & 71.91 & 37.56
& 51.05 & 72.85 & 80.55 & 37.81 \\
$M_3$ (STaG w/o $\mathcal{M}_v^t$)
& 43.44 & 64.42 & 72.87 & 38.56
& 51.60 & 74.40 & 81.20 & 38.43 \\
$M_4$ (STaG w/o ${g_t}$)
& 44.07 & 65.01 & 72.99 & 38.61
& 52.65 & 74.30 & 82.00 & 39.87 \\
$M_5$ (STaG w/o $\mathcal{L}_{CAE}$)
& 43.96 & 64.98 & 73.02 & 38.55
& 52.90 & 74.55 & 82.05 & 39.96 \\
\rowcolor[HTML]{D7F6FF}
\textbf{STaG}
& \textbf{44.23} & \textbf{65.09} & \textbf{73.08} & \textbf{38.64} & \textbf{53.70} & \textbf{75.00} & \textbf{82.60} & \textbf{40.15} \\
\hline\thickhline
\end{tabular}}
\vspace{-12pt}
\end{table}

\begin{wrapfigure}{r}{0.49\textwidth} 
  \vspace{-20pt}
  \centering
  \setlength{\abovecaptionskip}{4pt}
  \includegraphics[width=\linewidth]{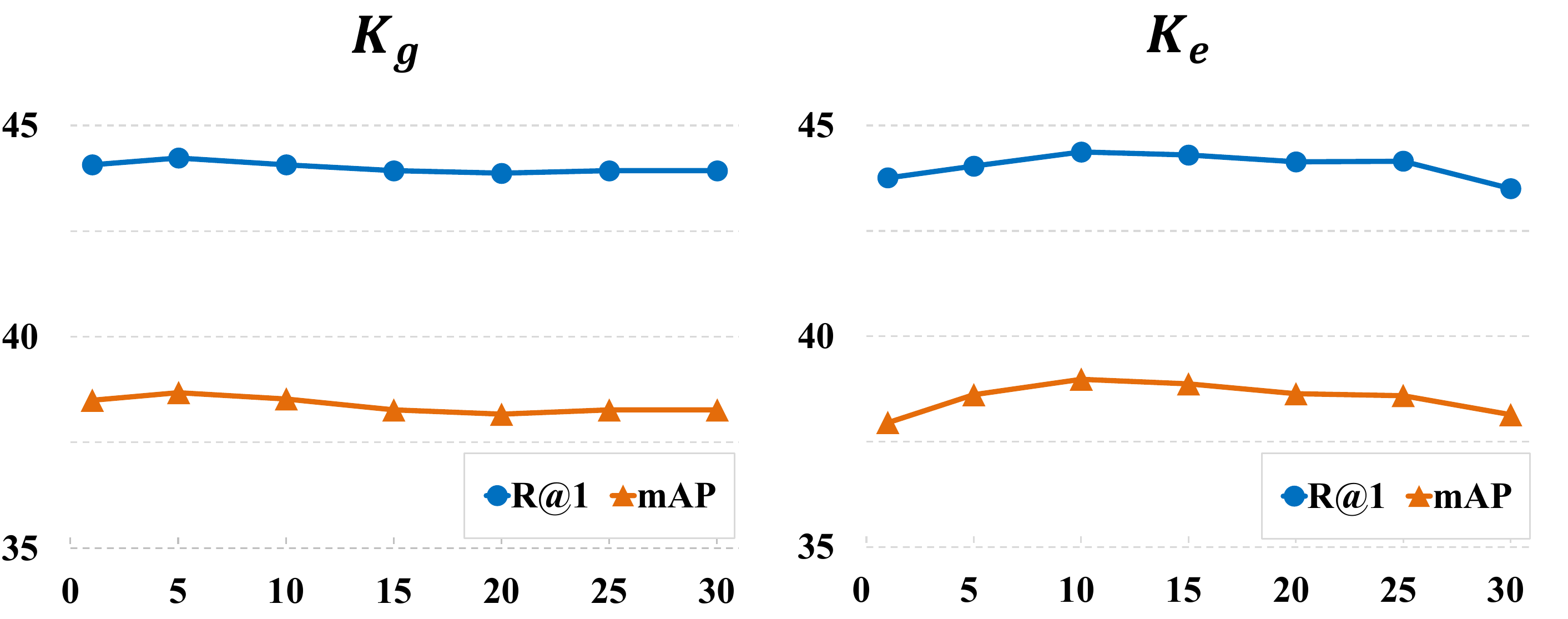}
  \caption{
    \textbf{Ablation of hyperparameters $K_g$ and $K_e$} using ICFG-PEDES as the source domain and CUHK-PEDES as target domain. (Left) Performance under varying neighbor sizes $K_g$ for graph construction in HGMA. (Right) Performance under varying top-$K$ sizes $K_e$ for the confidence-aware entropy regularization.} 
  \label{fig:hp-anaylsis}
  \vspace{-12pt}
\end{wrapfigure}

\vspace{-10pt}
\subsection{Hyper-parameter Analysis}
\label{sec: hyperparameters}
\vspace{-4pt}

HGMA has two hyperparameters: $K_g$ (Eq.~\ref{eq:topk_retrieval}), the number of visual neighbors retrieved per target sample, and $K_e$ (Eq.~\ref{eq:cae}), the number of source text neighbors used in $\mathcal{L}_{\mathrm{CAE}}$.
We vary one of the hyperparameters $(K_g, K_e)$ at a time while keeping the other fixed, and report the results in Fig.~\ref{fig:hp-anaylsis}.
It can be observed that both excessively small and excessively large values of $K_g$ or $K_e$ lead to suboptimal performance. A value that is too small provides insufficient evidence for propagation or confidence estimation, whereas a value that is too large introduces weakly related candidates.
We set $K_g{=}5$ and $K_e{=}10$ as the default values in our experiments.

\vspace{-12pt}
\section{Conclusion}
\vspace{-6pt}

In this paper, we formally address the highly practical yet under-explored setting of human-annotation-free UDA-TBPR. This paradigm poses significant challenges due to the complete
absence of human-written target-domain descriptions
and inherent visual domain shifts across datasets. To overcome these bottlenecks, we introduce STaG, a novel framework that synergistically reconstructs cross-modal semantics and mitigates visual domain gaps. Specifically, STaG leverages SATA to decouple identity-discriminative semantics from machine-generated stylistic artifacts, while employing HGMA to establish reliable cross-domain visual neighborhoods via heterogeneous graph consensus. Extensive experiments across multiple benchmarks demonstrate that STaG achieves state-of-the-art performance. Ultimately, by eliminating the reliance on target-domain human annotations, STaG paves the way for the scalable and cost-effective real-world deployment of person retrieval systems.

\vspace{-12pt}
\bibliography{references}
\bibliographystyle{ieeenat_fullname}
\vspace{-6pt}

\clearpage
\appendix
\section*{Appendix Overview}
\label{sec:appendix_overview}

To enhance the readability and navigation of this supplementary material, we outline the structural organization of the appendix as follows:

\noindent \textbf{A \hspace{0.2cm} Notation Summary} \dotfill \pageref{sec:notation} \\
\noindent \textbf{B \hspace{0.2cm} Prompt Template for SATA} \dotfill \pageref{app:prompt} \\
\noindent \textbf{C \hspace{0.2cm} Experimental Setup Details} \dotfill \pageref{app: exp setup} \\
\noindent \hspace*{0.5cm} C.1 \hspace{0.2cm} Datasets \dotfill \pageref{subsec:datasets} \\
\noindent \hspace*{0.5cm} C.2 \hspace{0.2cm} Protocol \dotfill \pageref{subsec:protocol} \\
\noindent \hspace*{0.5cm} C.3 \hspace{0.2cm} Implementation Details \dotfill \pageref{subsec:implementation} \\
\noindent \textbf{D \hspace{0.2cm} Detailed Mechanism of Style-Agnostic Text Alignment (SATA)} \dotfill \pageref{app:sata_details} \\
\noindent \textbf{E \hspace{0.2cm} Annotation Cost and Computational Overhead} \dotfill \pageref{sec:cost_overhead} \\
\noindent \hspace*{0.5cm} E.1 \hspace{0.2cm} Offline Generation Cost vs. Manual Annotation \dotfill \pageref{app:mllm costs} \\
\noindent \hspace*{0.5cm} E.2 \hspace{0.2cm} Training and Inference Complexity of STaG \dotfill \pageref{app: train complex} \\
\noindent \textbf{F \hspace{0.2cm} Pseudo-Text Reliability and Visual Robustness} \dotfill \pageref{sec:noise_robustness} \\
\noindent \hspace*{0.5cm} F.1 \hspace{0.2cm} Pseudo-Text Quality and Generation Robustness \dotfill \pageref{app:pseudotext_robustness} \\
\noindent \hspace*{0.5cm} F.2 \hspace{0.2cm} Robustness to Image Corruptions \dotfill \pageref{app:robustness_to_image_corruptions} \\
\noindent \textbf{G \hspace{0.2cm} Necessity of Semantic Alignment and Graph Topology} \dotfill \pageref{sec:sanity_checks} \\
\noindent \hspace*{0.5cm} G.1 \hspace{0.2cm} SATA: Quantifying Style Invariance \dotfill \pageref{app: distribution} \\
\noindent \hspace*{0.5cm} G.2 \hspace{0.2cm} SATA: Valid Semantics vs. Unstructured Noise \dotfill \pageref{subsec:valid_semantics} \\
\noindent \hspace*{0.5cm} G.3 \hspace{0.2cm} HGMA: KNN Topology vs. Random Neighbors \dotfill \pageref{subsec:knn_topology} \\
\noindent \hspace*{0.5cm} G.4 \hspace{0.2cm} Contribution Beyond Plain Pseudo-Text Adaptation \dotfill \pageref{app:plain_pseudotext} \\
\noindent \textbf{H \hspace{0.2cm} Additional Evaluations} \dotfill \pageref{app:appendix_experiments} \\
\noindent \hspace*{0.5cm} H.1 \hspace{0.2cm} Comparative Analysis with Human-Annotation-Accessible Baselines \dotfill \pageref{app:bridging_gap} \\
\noindent \hspace*{0.5cm} H.2 \hspace{0.2cm} Generalization Across Diverse Source Domains \dotfill \pageref{app:generalization} \\
\noindent \hspace*{0.5cm} H.3 \hspace{0.2cm} Architectural Robustness \dotfill \pageref{app:completing_matrix} \\
\noindent \hspace*{0.5cm} H.4 \hspace{0.2cm} Cross-Dataset Generalization of CAE \dotfill \pageref{app:cae_generalization} \\
\noindent \textbf{J \hspace{0.2cm} Visualization Analysis} \dotfill \pageref{app: visualization} \\
\noindent \hspace*{0.5cm} J.1 \hspace{0.2cm} Visualization of Image Activation Maps \dotfill \pageref{app:vis_image} \\
\noindent \hspace*{0.5cm} J.2 \hspace{0.2cm} Visualization of Attribute-Level Attention in Pseudo-Texts \dotfill \pageref{app:att in text} \\
\noindent \hspace*{0.5cm} J.3 \hspace{0.2cm} Visualization of Retrieved HGMA Neighbors \dotfill \pageref{app:hgma_neighbors} \\
\noindent \hspace*{0.5cm} J.4 \hspace{0.2cm} Visualization of Retrieval Results \dotfill \pageref{app:vis_retrieval}

\section{Notation Summary}
\label{sec:notation}

In Table~\ref{tab:notation_stag}, we present a comprehensive summary of the key notations and variables used in the formulation of our proposed STaG framework.

\begin{table}[htbp]
    \centering
    \setlength{\tabcolsep}{15pt}
    \caption{Summary of notations used in this paper, categorized by method components.}
    \label{tab:notation_stag}

    \renewcommand\arraystretch{1.3}
    \scalebox{0.9}{
    \begin{tabular}{l|l|c}
        \hline\thickhline
        \rowcolor{lightgray}
        \textbf{Symbol} & \multicolumn{1}{c|}{\textbf{Description}} & \textbf{Reference} \\
        \hline\hline

        $D^s, D^t$
        & Source and target domain datasets
        & Sec.~\ref{sec:Method} \\

        $v_i, t_i$
        & $\ell_2$-normalized image and text embeddings
        & Eq.~\ref{eq:def_vt} \\

        $\mathcal{L}_{itc}$
        & Symmetric image-text contrastive (InfoNCE) objective
        & Eq.~\ref{eq:clip-itc} \\

        $\hat{T}^s, \hat{T}^t$
        & Generator-synthesized pseudo-descriptions
        & Sec.~\ref{sec:sata} \\

        $\hat{\mathcal{D}}^s, \hat{\mathcal{D}}^t$
        & Augmented datasets with generated pairs
        & Sec.~\ref{sec:sata} \\

        $\mathcal{L}_{\mathrm{SATA}}$
        & Style-Agnostic Text Alignment optimization objective
        & Sec.~\ref{sec:sata} \\

        $\mathcal{M}_v^s, \mathcal{M}_t^s, \mathcal{M}_v^t$
        & FIFO Semantic Anchor Memories (SAMs)
        & Sec.~\ref{method: hgma} \\

        $\mathcal{N}_v^t$
        & Retrieved Top-$K_g$ intra-domain neighbor set
        & Eq.~\ref{eq:topk_retrieval} \\

        $\mathrm{Agg}(\cdot)$
        & Attention-based neighbor aggregation function
        & Eq.~\ref{eq:agg_def} \\

        $w_{t,n}^{(x)}$
        & Normalized attention weight for neighbor $n$
        & Eq.~\ref{eq:agg_def} \\

        $\hat{v}_t$
        & Fused target feature incorporating heterogeneous cues
        & Eq.~\ref{eq:hetero_fuse} \\

        $g_t$
        & Confidence-based residual gate for feature fusion
        & Sec.~\ref{method: hgma} \\

        $\tilde{v}_t$
        & Final corrected target image embedding
        & Sec.~\ref{method: hgma} \\

        $p_k$
        & Matching distribution over Top-$K_e$ source texts
        & Eq.~\ref{eq:cae} \\

        $\mathcal{L}_{\mathrm{CAE}}$
        & Confidence-Aware Entropy regularizer
        & Eq.~\ref{eq:cae} \\

        $\mathcal{L}$
        & Overall training objective for STaG
        & Eq.~\ref{eq:loss} \\

        \hline\thickhline
    \end{tabular}}
\end{table}

\section{Prompt Template for SATA}
\label{app:prompt}

Using the same generator and prompt encourages $\hat{T}^s$ and $\hat{T}^t$ to share similar generation styles, which facilitates learning style-invariant yet identity-discriminative representations rather than overfitting to domain-specific textual artifacts.
To encourage visual grounding and ensure descriptive fidelity, our prompt explicitly instructs the generator to describe solely observable attributes. Furthermore, we enforce identical decoding strategies for all inferences—specifically fixing the temperature $\tau$, nucleus sampling parameter $\text{top-}p$, and maximum sequence length $L$. The specific instruction used for generation is presented below:

\begin{center}
    \begin{tcolorbox}[
        colback=gray!5,
        colframe=gray!50,
        boxrule=1pt,
        arc=3pt,
        width=0.85\linewidth,
        left=6pt, right=6pt, top=6pt, bottom=6pt
    ]
        \textbf{\small System Instruction:} \\
        \small \fontfamily{cmtt}\selectfont\raggedright
        Write a description about the appearance of the person in the image.
        Do not imagine any contents that are not in the image.
    \end{tcolorbox}
\end{center}

\section{Experimental Setup Details}
\label{app: exp setup}
\subsection{Datasets}
\label{subsec:datasets}
Our experiments involve three open benchmark datasets.

\textbf{CUHK-PEDES}~\citep{li2017person} is the most commonly-used dataset in TBPS. It consists of $40,206$ images and $80,440$ texts from $13,003$ identities in total, which are split into $34,054$ images and $68,126$ texts from $11,003$ identities in the training set, $3,078$ images and $6,158$ texts from $1,000$ identities in the validation set, and $3,074$ images and $6,156$ texts from $1,000$ identities in the test set. The average length of all texts is $23$.

\textbf{ICFG-PEDES}~\citep{ding2021semantically} contains $54,522$ images from $4,102$ identities. Each of the images is described by one text. The dataset is split into $34,674$ images from $3,102$ identities in the training set, and $19,848$ images from $1,000$ identities in the test set. There are $37$ words per text on average.

\textbf{RSTPReid}~\citep{zhu2021dssl} consists of $20,505$ images of $4,101$ identities. Each identity has $5$ corresponding images captured from different cameras. Each image is annotated with $2$ textual descriptions, and each description is no shorter than $23$ words. There are $3,701$/$200$/$200$ identities utilized for training/validation/testing, respectively.

\subsection{Protocol}
\label{subsec:protocol}
We adopt the widely-used Rank@K (R@K for short, K=1, 5, 10) metric to evaluate the performance of the proposed
framework. Specifically, given a query text, we rank all the test images via the similarity with the query text, and the search is deemed to be successful if top-K images contain any corresponding identity. R@K is the percentage of successful searches. In addition, we also adopt the mean average precision (mAP) as a complementary metric. Rank@K reflects the accuracy of the first few retrieval results, while mAP emphasizes the comprehensive performance of the method.

\subsection{Implementation Details}
\label{subsec:implementation}
During training, all images are resized to $224 \times 224$, and each text input is truncated to at most 77 tokens.
We optimize the retrieval model using AdamW with an initial learning rate of $1\times 10^{-5}$ and a cosine learning-rate scheduler, and train for $5$ epochs with a batch size of $128$.
We adopt InternVL2.5-8B~\citep{chen2024expanding} as the generator.
The hyperparameters are set as follows.
We set the numbers of neighbors $K_g=5$ and the number of entropy candidates $K_e=10$ (in Section~\ref{method: hgma}).
The temperature parameters in Eq.~\ref{eq:cae} is set to $\tau_e=0.2$.
The capacity of SAMs in Section~\ref{method: hgma} is set to $2^{14}$ .
All experiments are conducted on 4 NVIDIA L40 GPUs.

\section{Detailed Mechanism of Style-Agnostic Text Alignment (SATA)}
\label{app:sata_details}

\begin{figure}[t]
  \centering
  \setlength{\abovecaptionskip}{4pt}

  \includegraphics[width=0.9\linewidth]{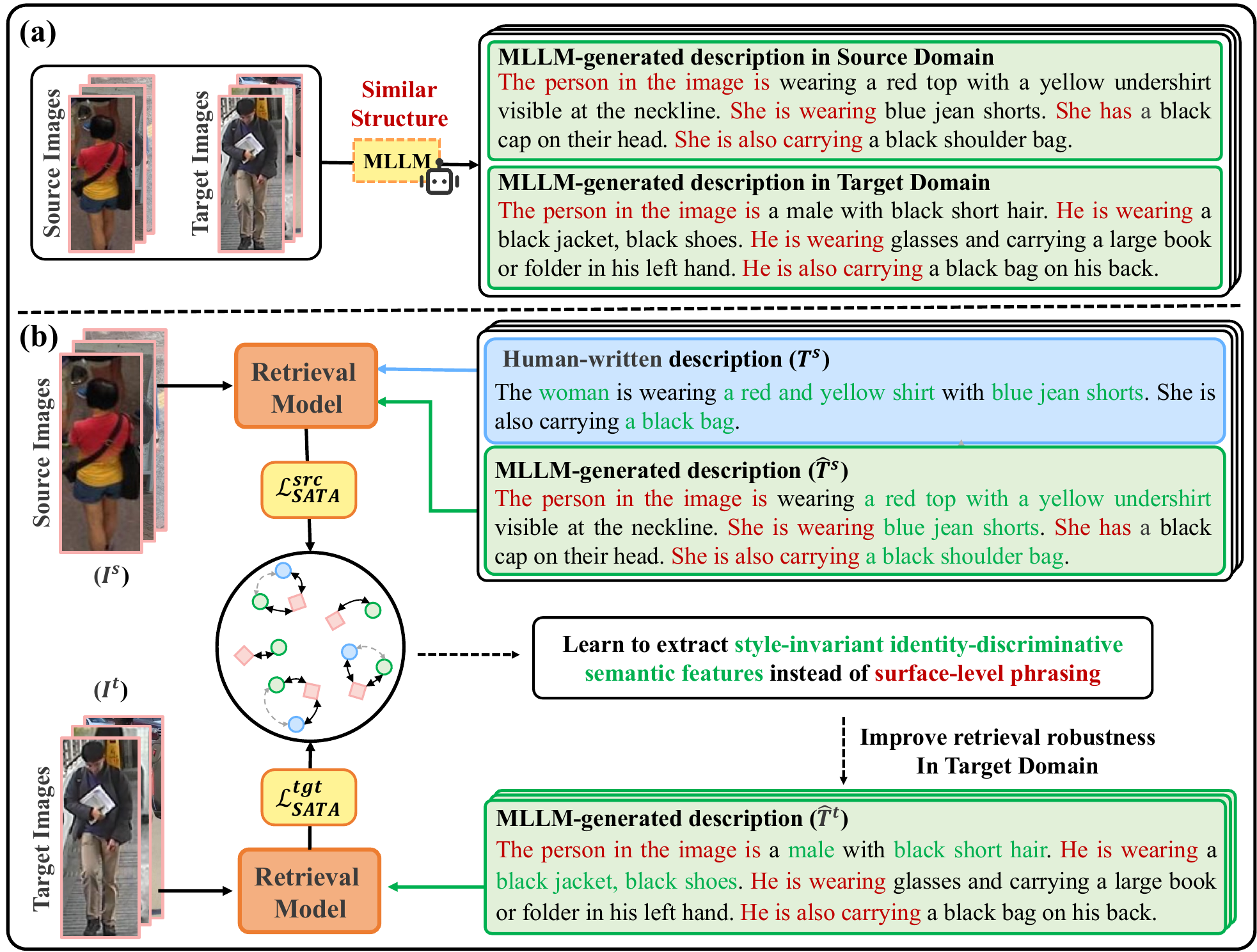}
  \caption{\textbf{Detailed mechanism of the Style-Agnostic Text Alignment (SATA) module.} (a) Given the identical MLLM generator and prompt, the synthesized pseudo-descriptions in both domains exhibit highly similar, rigid syntactic structures (highlighted in red), which severely diverge from the flexible phrasing of human-written queries.  (b)  illustrates the fundamental stylistic divergence between human-written queries and MLLM-generated pseudo-descriptions. By utilizing a shared MLLM generator, machine-generated artifacts remain structurally consistent across domains. SATA forces the retrieval model to align both text modalities to a shared visual anchor in the source domain $(\mathcal{L}_{SATA}^{Src})$, compelling the network to discard rigid syntactic templates (highlighted in red) and extract pure identity-discriminative semantics (highlighted in green). This calibrated capability seamlessly generalizes to the target domain $(\mathcal{L}_{SATA}^{tgt})$, ensuring robust adaptation.}
  \label{fig:app_sata}
\end{figure}

To provide a deeper mechanistic understanding of how SATA effectively bridges the cross-modal domain gap, Fig.~\ref{fig:app_sata} visually dissects the text generation and alignment pipeline. Specifically, we elucidate the module's operation across three critical dimensions:

\ding{182} The Stylistic Divergence and Shared Artifacts: As depicted in the text boxes of Fig.~\ref{fig:app_sata}, human-written descriptions in the source domain ($T^s$) typically exhibit flexible phrasing and focus directly on discriminative attributes, such as "The woman is wearing a red and yellow shirt...". In contrast, texts generated by the MLLM, denoted as $\hat{T}^s$ and $\hat{T}^t$, suffer from rigid, repetitive stylistic artifacts. These include introductory boilerplate like "The person in the image is..." and structured enumerations like "She is wearing..." or "She has...". This severe stylistic divergence is precisely why relying solely on target-domain pseudo-texts for cross-modal alignment yields sub-optimal performance, a vulnerability empirically validated by the limited gains of variant $M_2$ in our ablation study shown in Table~\ref{tab:ablation_sata}. Overcoming this fundamental bottleneck is the core motivation behind SATA. In Fig.~\ref{fig:app_sata}, we highlight these uninformative, MLLM-specific syntactic templates in red, while the actual identity-discriminative semantic concepts (clothing colors and accessories), are highlighted in~green.

\ding{183} Source Domain: Calibrating the Semantic Filter via $\mathcal{L}_{SATA}^{Src}$. The left portion of Fig.~\ref{fig:app_sata} illustrates the core calibration phase. In the source domain, a single image anchor $I^s$ is paired with both its original human annotation $T^s$ and its MLLM-generated counterpart $\hat{T}^s$. When optimizing the contrastive objective $\mathcal{L}_{SATA}^{Src}$, the Retrieval Model is mathematically forced to pull these two stylistically distinct texts toward the identical visual point in the shared embedding space. To minimize this loss, the text encoder has no choice but to act as a semantic filter: it must actively discard the differing surface-level phrasing highlighted in red, and isolate the shared underlying visual attributes highlighted in green. This process explicitly decouples the style-invariant semantic features from the machine-generated~noise.

\ding{184} Target Domain: Robust Cross-Modal Adaptation via $\mathcal{L}_{SATA}^{tgt}$. The crucial bridge between the source and target domains is denoted by the "Similar Structure" arrow in Fig.~\ref{fig:app_sata}(a). The pseudo-descriptions for both the source $\hat{T}^s$ and target $\hat{T}^t$ domains are synthesized utilizing the identical frozen MLLM and prompt template, they inherently share the exact same rigid syntactic structures, referred to as the red artifacts. Consequently, when the target objective $\mathcal{L}_{SATA}^{tgt}$ aligns the unlabeled target images $I^t$ with $\hat{T}^t$, the text encoder has already been pre-calibrated by the source domain to ignore these specific formatting artifacts. Instead of overfitting to the MLLM's phrasing, the model robustly extracts the pure identity semantics, successfully mitigating the cross-domain shift and establishing highly reliable image-text alignment in the human-annotation-free target domain.

\section{Annotation Cost and Computational Overhead}
\label{sec:cost_overhead}
As highlighted in our motivation (Sec.~\ref{sec:intro}), the primary bottleneck for real-world UDA-TBPR deployment is the prohibitive cost of acquiring target-domain textual annotations. To demonstrate that introducing an MLLM is a genuine cost-reduction paradigm, rather than simply replacing human labor costs with expensive computational overhead, we provide a comprehensive analysis of the offline preparation expenses and the training complexity of our framework.

\subsection{Offline Generation Cost vs. Manual Annotation}
\label{app:mllm costs}

\textbf{\ding{182} Economic Cost and Efficiency.}
Manual annotation for TBPR is notoriously labor-intensive and expensive. On typical crowdsourcing platforms (e.g., Amazon Mechanical Turk), accurately annotating a single pedestrian image requires 1 to 2 minutes. At a standard minimum wage of \$6.00/hour, the manual annotation cost equates to approximately \$0.10 per image.
In contrast, our offline generation using InternVL2.5-8B on four NVIDIA L40 GPUs processes images at approximately $0.11$ seconds per image.
Under a system-level compute price of approximately \$4.00 per four-GPU hour, this corresponds to about \$0.00012 per image, yielding roughly an $800\times$ reduction compared with the estimated \$0.10 manual annotation cost.

\textbf{\ding{183} Scalability for Large-Scale Deployment.}
Beyond direct monetary costs, MLLM generation scales deterministically and is exempt from human-centric bottlenecks such as annotator fatigue, subjective inconsistency, and logistical coordination overhead.

For instance, manually annotating a new target domain of 100,000 surveillance images would cost roughly \$10,000 and require months to complete. Our framework accomplishes the identical task in approximately 3 hours for about \$12 in compute costs. The estimated energy footprint for this process (based on the 300W TDP of an L40 GPU) is around $3.67$ kWh, which is a negligible expenditure given the magnitude of the data processed.

\textbf{\ding{184} Zero Inference Overhead.}
Crucially, the MLLM is employed exclusively as an offline data preparation engine. Once the pseudo-texts are generated and the model is trained, the MLLM is entirely discarded. During online deployment and inference, only the lightweight CLIP-style dual-encoder is active. Thus, STaG introduces zero architectural changes, latency delays, or deployment complexities to the final retrieval system. Trading a one-time, marginal offline computational cost for substantial cross-domain performance improvements represents a highly practical and favorable trade-off for real-world applications.

\subsection{Training and Inference Complexity of STaG}
\label{app: train complex}
We analyze both the asymptotic complexity and the measured computational overhead of STaG.
Let $B$ denote the mini-batch size, $D$ the embedding dimension, and $M$ the fixed capacity of each Semantic Anchor Memory.

\textbf{\ding{182} SATA Training Complexity.}
SATA introduces no additional trainable network module beyond the underlying retrieval backbone.
Its main additional computation arises from image--text contrastive optimization over generated pairs.
Computing a dense pairwise similarity matrix for a batch of $B$ image and text embeddings of dimension $D$ requires $\mathcal{O}(B^2D)$ operations.
The pseudo-descriptions themselves are generated offline and remain fixed throughout retrieval-model training.

\textbf{\ding{183} HGMA Training Complexity.}
The Heterogeneous Graph Memory Alignment (HGMA) module introduces structural alignment. It maintains three fixed memory queues ($\mathcal{M}_{v}^{s}, \mathcal{M}_{t}^{s}, \mathcal{M}_{v}^{t}$), each with a capacity of $M$ (set to 16,384) and a dimension of $D$ (set to 512).
The static memory complexity is \(O(MD)\) for each memory bank, and remains independent of the overall dataset size. The computational bottleneck is nearest-neighbor retrieval across the memory banks. Computing similarities between a batch of \(B\) query features and \(M\) memory features of dimension \(D\) requires \(O(BMD)\) operations. Since \(M\) is fixed, the memory and retrieval costs remain bounded with respect to the dataset size.
Crucially, since $M$ is a predefined constant, this similarity computation is strictly restricted to a local subset, completely decoupling the graph construction overhead from the overall target dataset scale.

\textbf{\ding{184} Measured Training and Memory Overhead.}
Under the ICFG-PEDES $\rightarrow$ CUHK-PEDES setting, a complete training run takes approximately 25 minutes on four NVIDIA L40 GPUs, with approximately 40 GB peak memory usage measured on a single L40 GPU.
STaG introduces 1.05M additional trainable parameters, while the three Semantic Anchor Memories require approximately 96 MiB in FP32 of storage in total.
For inference, we measure a single-query latency of 5.78 ms and a throughput of 373 queries/s with a batch size of 64.

\textbf{\ding{185} Online Inference Complexity.}
Most importantly, the added complexity of STaG is strictly confined to the training phase. Once the model is fully trained, the MLLM, the Semantic Anchor Memories (SAMs), and the graph aggregation operations are entirely discarded. During online deployment and inference, only the lightweight CLIP-style dual-encoder is active. Thus, STaG introduces zero architectural changes, latency delays, or deployment complexities to the final retrieval system.

\begin{table}[t]
\centering
\caption{Robustness to different MLLM generators (ICFG-PEDES $\rightarrow$ CUHK-PEDES). Retrieval performance remains stable regardless of which MLLM is used for offline pseudo-text generation.}
\label{tab:generator_robustness}
\setlength\tabcolsep{12pt}
\renewcommand\arraystretch{1.2}
\begin{tabular}{lcccc}
\hline\thickhline
Generator Model & R@1 & R@5 & R@10 & mAP \\
\midrule
InternVL2.5-8B (Default) & 44.23 & 65.09 & 73.08 & 38.64 \\
InternVL2-8B & 44.07 & 64.31 & 71.99 & 38.40 \\
Qwen2-VL-7B-Instruct & 43.84 & 64.11 & 71.15 & 37.94 \\
\hline\thickhline
\end{tabular}
\end{table}

\begin{table}[t]
\centering
\caption{Robustness to hallucination noise (ICFG-PEDES $\rightarrow$ CUHK-PEDES). We randomly alter observable attributes in the target pseudo-texts to simulate varying degrees of MLLM hallucinations.}
\label{tab:noise_robustness}
\setlength\tabcolsep{12pt}
\renewcommand\arraystretch{1.2}
\begin{tabular}{lcccc}
\hline\thickhline
Noise Rate & R@1 & R@5 & R@10 & mAP \\
\midrule
0\% (Original Pseudo-texts) & 44.23 & 65.09 & 73.08 & 38.64 \\
10\% & 43.87 & 64.55 & 72.63 & 38.28 \\
20\% & 44.05 & 64.88 & 72.99 & 38.49 \\
30\% & 43.13 & 62.85 & 71.23 & 37.67 \\
\hline\thickhline
\end{tabular}
\end{table}

\section{Pseudo-Text Reliability and Visual Robustness}
\label{sec:noise_robustness}

\subsection{Pseudo-Text Quality and Generation Robustness}
\label{app:pseudotext_robustness}

SATA relies on pseudo-descriptions generated by an off-the-shelf MLLM.
We therefore examine their reliability from three aspects: intrinsic quality, sensitivity to the generator choice, and robustness to hallucinated attributes.

\textbf{\ding{182} Quality of Generated Pseudo-Texts.}
We compare the pseudo-descriptions generated by InternVL2.5-8B with human annotations on CUHK-PEDES.
They achieve a CLIP cosine similarity of 73.89, a ROUGE-L score of 24.79, and BLEU-1/2/4 scores of 28.01/17.23/6.68.
The 97.39\% unique-caption ratio further indicates that the generated descriptions retain substantial lexical diversity.
These results support their use as target-domain cross-modal supervision.

\textbf{\ding{183} Robustness to the Choice of MLLM Generator.}
Table~\ref{tab:generator_robustness} compares InternVL2.5-8B with InternVL2-8B and Qwen2-VL-7B-Instruct on ICFG-PEDES $\rightarrow$ CUHK-PEDES.
Rank@1 varies only from 43.84\% to 44.23\%, showing that STaG is relatively insensitive to the choice of generator among the evaluated MLLMs.

\textbf{\ding{184} Robustness to Hallucination Noise.}
To evaluate tolerance to factual errors, we randomly corrupt $r\%$ of the target pseudo-descriptions by replacing observable attributes, such as clothing colors, garment types, and hairstyles, with plausible but incorrect alternatives.
Here, $r$ denotes the proportion of corrupted descriptions rather than the token-level corruption ratio.
As shown in Table~\ref{tab:noise_robustness}, Rank@1 decreases only from 44.23\% to 43.13\% when 30\% of the pseudo-descriptions are corrupted.

This robustness benefits from two complementary mechanisms.
SATA anchors the representation space with clean source-domain annotations, while HGMA aggregates multiple source-side semantic references, reducing the influence of individual noisy pseudo-descriptions.

\subsection{Robustness to Image Corruptions}
\label{app:robustness_to_image_corruptions}

We further evaluate the robustness of STaG to visual degradation at test time.
Specifically, we apply five common corruptions (Gaussian blur, motion blur, low illumination with noise, partial occlusion, and JPEG compression) to the target-domain test images under ICFG-PEDES $\rightarrow$ CUHK-PEDES.

As shown in Table~\ref{tab:image_corruption_robustness}, STaG exhibits only a $0.19$-point average Rank@1 drop under mild corruptions, compared with $1.27$ points for the baseline.
Under severe corruptions, both methods degrade, but STaG consistently remains above the baseline, achieving an average Rank@1 of $37.65$ versus $28.87$.

These results show that the adaptation learned by STaG remains effective when target-domain gallery images suffer from common test-time degradations.
Since the corruptions are introduced only at evaluation time, this experiment specifically measures retrieval robustness to degraded visual inputs rather than the robustness of MLLM pseudo-text generation.

\begin{table}[t]
\centering
\caption{
\textbf{Robustness to target-domain test-image corruptions}
on ICFG-PEDES $\rightarrow$ CUHK-PEDES.
$\Delta$ denotes the R@1 change relative to clean test images.
}
\label{tab:image_corruption_robustness}
\small
\setlength{\tabcolsep}{5pt}
\renewcommand{\arraystretch}{1.15}
\begin{tabular}{llrrrr}
\hline\thickhline
Corruption & Parameters &
\multicolumn{2}{c}{STaG} &
\multicolumn{2}{c}{Baseline} \\
\cmidrule(lr){3-4}\cmidrule(lr){5-6}
 & & R@1 & $\Delta$ & R@1 & $\Delta$ \\
\midrule
Original images & -- & 44.23 & -- & 35.99 & -- \\
\midrule
\multicolumn{6}{l}{\textit{Mild corruption}} \\
Gaussian blur
& $\sigma=1$ & 44.39 & +0.16 & 35.77 & -0.22 \\
Motion blur
& $k=7$ & 44.57 & +0.34 & 35.86 & -0.13 \\
Low light + noise
& $\gamma=1.5,\ \sigma_n\approx0.008$ & 44.61 & +0.38 & 34.70 & -1.29 \\
Occlusion
& 10\% image area & 43.30 & -0.93 & 32.75 & -3.24 \\
JPEG compression
& $q=50$ & 43.32 & -0.91 & 34.51 & -1.48 \\
\textit{Average}
& -- & 44.04 & -0.19 & 34.72 & -1.27 \\
\midrule
\multicolumn{6}{l}{\textit{Severe corruption}} \\
Gaussian blur
& $\sigma=3$ & 36.04 & -8.19 & 28.11 & -7.88 \\
Motion blur
& $k=15$ & 41.07 & -3.16 & 32.56 & -3.43 \\
Low light + noise
& $\gamma=2.5,\ \sigma_n\approx0.024$ & 35.99 & -8.24 & 25.77 & -10.22 \\
Occlusion
& 25\% image area & 35.52 & -8.71 & 26.81 & -9.18 \\
JPEG compression
& $q=20$ & 39.61 & -4.62 & 31.08 & -4.91 \\
\textit{Average}
& -- & 37.65 & -6.58 & 28.87 & -7.12 \\
\hline\thickhline
\end{tabular}
\end{table}

\section{Necessity of Semantic Alignment and Graph Topology}
\label{sec:sanity_checks}

To further examine the roles of SATA and HGMA, we provide several complementary sanity checks and control experiments on the ICFG-PEDES $\rightarrow$ CUHK-PEDES setting.
These analyses evaluate style alignment, semantic validity, neighborhood topology, and the contribution beyond plain pseudo-text adaptation.

\subsection{SATA: Quantifying Style Invariance.}
\label{app: distribution}
\begin{figure}[t]
\centering
\includegraphics[width=0.9\linewidth]{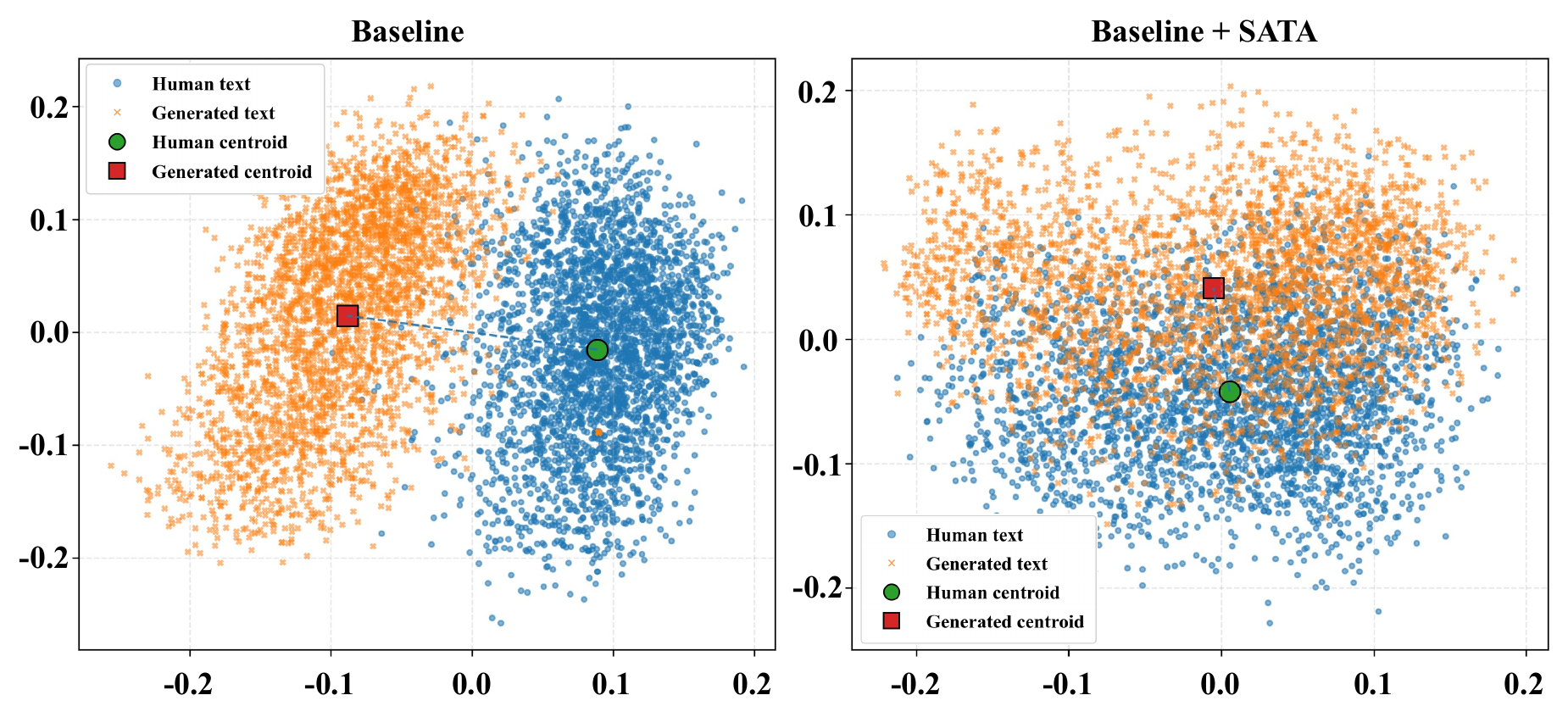}
\caption{Visualization of feature distributions for human-written texts (blue dots) and MLLM-generated pseudo-texts (orange crosses). \textbf{Left:} The baseline model exhibits a distinct stylistic domain gap, with the two text modalities forming disjoint clusters. \textbf{Right:} By integrating SATA, the two distributions are pulled into a highly overlapping and unified semantic space, significantly reducing the distance between their centroids (green circle and red square).}
\label{fig:style_distribution}
\end{figure}

To directly validate that SATA successfully bridges the stylistic domain gap between human-written queries and machine-generated descriptions, we evaluate the feature embeddings of both textual modalities on the source domain. As summarized in Table~\ref{tab:style_invariance} and visually corroborated by Fig.~\ref{fig:style_distribution}, SATA aligns the representations across two critical levels:

\ding{182}  \textbf{Macroscopic Distribution Alignment:} As illustrated in Fig.~\ref{fig:style_distribution}, the baseline model exhibits two disjoint clusters for human and generated texts, indicating a severe stylistic gap. SATA elegantly maps these distinct styles into a unified semantic space, merging the initially separated distributions and reducing their Centroid L2 Distance by more than half ($0.2821 \rightarrow 0.1275$).

\ding{183} \textbf{Microscopic Instance-Level Alignment:} At the instance level, SATA significantly boosts the Pairwise Cosine Mean between human queries and their corresponding pseudo-texts by $+0.1109$.
The proportion of highly similar human--generated text pairs, measured by $P(\cos > 0.90)$, increases by an absolute $13.55\%$.

Together, these measurements provide evidence that SATA reduces the representation gap between human-written and generated descriptions while preserving their shared semantic content. Token-level attention visualization can be seen in Appendix~\ref{app: visualization}.

\begin{table}[t]
\centering
\caption{Quantifying style invariance (Human vs. Generated Texts). We compare the feature embeddings of human annotations and MLLM-generated pseudo-texts on the source domain.}
\label{tab:style_invariance}
\setlength\tabcolsep{5pt}
\renewcommand\arraystretch{1.15}
\begin{tabular}{lccc}
\hline\thickhline
Metric & Baseline & Baseline + SATA (Ours) & Change \\
\midrule
Centroid L2 Distance $\downarrow$ & 0.2821 & 0.1275 & $\Delta = -0.1546$ \\
Pair Cosine Mean $\uparrow$ & 0.7883 & 0.8992 & $\Delta = +0.1109$ \\
$P(\cos > 0.90)$ $\uparrow$ & 37.80\% & 51.35\% & $\Delta = +13.55\%$ \\
\hline\thickhline
\end{tabular}
\end{table}

\subsection{SATA: Valid Semantics vs. Unstructured Noise.}
\label{subsec:valid_semantics}
We first investigate whether the SATA module relies on meaningful visual-semantic grounding or if it merely acts as a generic text-based regularizer. To this end, we replace the valid generated pseudo-descriptions with randomly sampled noise. Concretely, we randomly select 30\% of the target samples and, within each selected sample, replace the attribute tokens with words drawn uniformly at random from the vocabulary; the remaining 70\% retain their original pseudo-texts. This matches the sample-level selection used in the hallucination experiment, but substitutes random tokens rather than plausible but incorrect ones.

As shown in Table~\ref{tab:sata_sanity}, random token corruption yields 38.50\% Rank@1, notably below the 43.13\% obtained with structured hallucination noise at the same 30\% sample corruption rate.
This gap indicates that SATA benefits from meaningful visual semantic content rather than from the mere presence of an auxiliary text signal.
At the same time, the smaller degradation under structured hallucinations suggests that the framework tolerates moderate attribute-level errors better than unstructured semantic corruption.

\begin{table}[t]
\centering
\caption{Sanity check on the SATA module (ICFG-PEDES $\rightarrow$ CUHK-PEDES). We compare the impact of introducing unstructured random noise versus structured MLLM hallucinations.}
\label{tab:sata_sanity}
\setlength\tabcolsep{10pt}
\renewcommand\arraystretch{1.2}
\begin{tabular}{lcccc}
\hline\thickhline
Method & R@1 & R@5 & R@10 & mAP \\
\midrule
Baseline (HGMA Only) & 36.63 & 58.08 & 67.46 & 32.94 \\
+ Random Noise Text & 38.50 & 60.00 & 68.47 & 33.24 \\
+ 30\% Hallucination & 43.13 & 62.85 & 71.23 & 37.67 \\
+ SATA (Original) & \textbf{44.23} & \textbf{65.09} & \textbf{73.08} & \textbf{38.64} \\
\hline\thickhline
\end{tabular}
\end{table}

\begin{table}[t]
\centering
\caption{Sanity check on the HGMA module (ICFG-PEDES $\rightarrow$ CUHK-PEDES). We compare random neighbor sampling against our proposed KNN-based graph retrieval.}
\label{tab:hgma_sanity}
\setlength\tabcolsep{10pt}
\renewcommand\arraystretch{1.2}
\begin{tabular}{lcccc}
\hline\thickhline
Method & R@1 & R@5 & R@10 & mAP \\
\midrule
Baseline (SATA Only) & 41.03 & 59.71 & 67.46 & 36.11 \\
+ Random Neighbor & 41.57 & 60.29 & 67.88 & 36.40 \\
+ HGMA (KNN-based) & \textbf{44.23} & \textbf{65.09} & \textbf{73.08} & \textbf{38.64} \\
\hline\thickhline
\end{tabular}
\end{table}

\begin{table}[t]
\centering
\caption{
\textbf{Comparison with plain pseudo-text adaptation}
on ICFG-PEDES $\rightarrow$ CUHK-PEDES.
}
\label{tab:plain_pseudotext}
\small
\setlength{\tabcolsep}{10pt}
\renewcommand{\arraystretch}{1.15}
\begin{tabular}{lcccc}
\hline\thickhline
Method & R@1 & R@5 & R@10 & mAP \\
\midrule
Baseline
& 35.99 & 56.94 & 66.23 & 32.07 \\
Plain pseudo-texts
& 38.85 & 59.07 & 67.28 & 34.79 \\
\rowcolor[HTML]{D7F6FF}
\textbf{STaG}
& \textbf{44.23} & \textbf{65.09} & \textbf{73.08} & \textbf{38.64} \\
\hline\thickhline
\end{tabular}
\end{table}

\subsection{HGMA: KNN Topology vs. Random Neighbors.}
\label{subsec:knn_topology}
We apply the analogous test to HGMA by asking whether its gains arise from the structure of the retrieved neighborhood or merely from the addition of memory-based features. We replace the $K$-Nearest Neighbor ($K_g$) retrieval mechanism in Eq.~\ref{eq:topk_retrieval} with uniform random sampling from the semantic anchor memories, keeping memory capacity, attention aggregation, residual gating, and the CAE regularizer unchanged.
Table~\ref{tab:hgma_sanity} shows that random-neighbor aggregation yields only +0.54\% Rank@1 over the SATA-only baseline (41.03\% $\rightarrow$ 41.57\%), whereas KNN-based HGMA yields +3.20\% (41.03\% $\rightarrow$ 44.23\%).
The gap indicates that HGMA benefits from similarity-based neighborhood construction rather than generic feature aggregation.
Retrieving feature-similar neighbors provides more coherent cross-domain references than random sampling, without assuming any cross-domain identity correspondence.

\subsection{Contribution Beyond Plain Pseudo-Text Adaptation}
\label{app:plain_pseudotext}

To isolate the contribution of STaG from that of MLLM-generated supervision itself, we additionally evaluate a plain pseudo-text baseline that directly adapts the retrieval model using the same target image--pseudo-text pairs, without SATA or HGMA.

As shown in Table~\ref{tab:plain_pseudotext}, pseudo-text supervision alone improves Rank@1 from $35.99$ to $38.85$, confirming that the generated descriptions provide useful target-domain supervision.
STaG further increases Rank@1 to $44.23$ and mAP to $38.64$, outperforming the plain pseudo-text baseline by $5.38$ and $3.85$ points, respectively.
Together with the component ablations in the main paper, this result shows that the improvement of STaG extends beyond the benefit of raw pseudo-text supervision.

\section{Additional Evaluations}
\label{app:appendix_experiments}

This section provides additional evaluations of STaG from several complementary perspectives, including comparisons across different supervision settings, generalization across source domains, architectural compatibility, cross-dataset behavior of CAE, and run-to-run stability.
These experiments supplement the controlled comparisons reported in the main paper.

\begin{table*}[t]
\centering
\caption{
Comparisons with state-of-the-art methods using \textbf{CUHK-PEDES} as the source domain and \textbf{ICFG-PEDES / RSTPReid} as target domains.
We report results under three target-domain supervision settings:
(1) \textbf{Fully Supervised}: uses ground-truth paired image--text annotations from the target domain, serving as a reference under full target-domain supervision;
(2) \textbf{Text-Accessible UDA-TBPR}: uses human-written target-domain descriptions during adaptation without requiring paired target image--text annotations; and
(3) \textbf{Human-Annotation-Free UDA-TBPR (Ours)}: uses unlabeled target images and MLLM-generated pseudo-descriptions, without accessing human-written target-domain descriptions during training or adaptation.
Results across different supervision settings are reported as cross-setting references rather than strictly controlled comparisons.
``Backbone+STaG'' further evaluates the compatibility of STaG with different retrieval architectures under the human-annotation-free setting.
Controlled analyses under matched human-annotation-free conditions are provided in Sec.~\ref{sec:ablation} and Appendix~\ref{app:plain_pseudotext}.
* indicates results reproduced by us due to the unavailability of official source code.
}
\label{tab:c2all}
\small{
\resizebox{1\linewidth}{!}{
\setlength\tabcolsep{6pt}
\renewcommand\arraystretch{1.4}
    \begin{tabular}{cllcccccccc}
    \hline\thickhline
     &  &  & \multicolumn{4}{c}{CUHK-PEDES $\rightarrow$ ICFG-PEDES} & \multicolumn{4}{c}{CUHK-PEDES $\rightarrow$ RSTPReid} \\
    \cmidrule(lr){4-7} \cmidrule(lr){8-11}
    \multirow{-2}{*}{Setting} & \multirow{-2}{*}{Method} & \multirow{-2}{*}{Ref.} & R@1 & R@5 & R@10 & mAP & R@1 & R@5 & R@10 & mAP \\
    \hline\hline
    \multirow{7}{*}{Fully Supervised}
    & IRRA~\citep{cvpr23crossmodal} & CVPR'23
    & 63.46 & 80.25 & 85.82 & 38.06
    & 60.20 & 81.30 & 88.20 & 47.17 \\
    & RaSa~\citep{bai2023rasa} & IJCAI'23
    & 65.28 & 80.40 & 85.12 & 41.29
    & 66.90 & 86.50 & 91.35 & 52.31  \\
    & TBPS-CLIP~\citep{cao2024empirical} & AAAI'24
    & 65.05 & 80.34 & 85.47 & 39.83
    & 61.60 & 81.20 & 87.70 & 47.71 \\
    & RDE~\citep{qin2024noisy} & CVPR'24
    & 67.68 & 82.47 & 87.36 & 40.06
    & 65.35 & 83.95 & 89.90 & 50.88 \\
    & PTMI~\citep{lu2025prompt} & TIFS'25
    & 66.54 & 81.50 & 86.50 & 46.05
    & 63.35 & 81.95 & 89.10 & 52.46 \\
    & GAHR~\citep{qi2025granularity} & TIFS'25
    & 68.69 & 82.83 & 87.40 & 42.10
    & 68.85 & 86.50 & 91.10 & 53.60 \\
    & ICL~\citep{qin2025human} & CVPR'25
    & 69.22 & 83.49 & 88.06 & 42.34
    & 70.00 & 86.60 & 91.70 & 54.16 \\
    \hline
    \multirow{3}{*}{\makecell[c]{UDA-TBPR}}
    & POUF~\citep{Tanwisuth2023POUFPU}   & ICML'23
    & 41.08 & 61.13 & 69.02 & 19.06
    & 43.04 & 64.10 & 71.09 & 21.05 \\
    & ReCLIP~\citep{Hu2023ReCLIPRC} & WACV'24
    & 48.17 & 69.08 & 75.22 & 26.04
    & 52.15 & 76.20 & 82.16 & 37.10\\
    & GCKD*~\citep{Luo2025GraphBasedCK}   & AAAI'25
    & 51.16 & 71.45 & 78.50 & 31.34
    & 57.44 & 78.50 & 84.70 & 40.12 \\
    \hline
    \rowcolor[HTML]{D7F6FF}
    {\raisebox{-0.5\height}{\shortstack[c]{\strut Human-Annotation-Free \\ \strut UDA-TBPR}}}
    & RaSa+STaG & Ours
    & 50.45 & 67.24 & 73.55 & 24.23
    & 60.75 & 82.25 & 87.15 & 45.88 \\
    \cline{2-11}
    \hline\thickhline
\end{tabular}
}}

\end{table*}

\begin{table*}[t]
\centering
\caption{
Comparisons with state-of-the-art methods using \textbf{ICFG-PEDES} as the source domain and \textbf{CUHK-PEDES / RSTPReid} as target domains.
We follow the same target-domain supervision settings and cross-setting reference protocol defined in Table~\ref{tab:c2all}.
}
\label{app:i2all}
\small{
\resizebox{1\linewidth}{!}{
\setlength\tabcolsep{6pt}
\renewcommand\arraystretch{1.4}
    \begin{tabular}{cllcccccccc}
    \hline\thickhline
     &  &  & \multicolumn{4}{c}{ICFG-PEDES $\rightarrow$ CUHK-PEDES} & \multicolumn{4}{c}{ICFG-PEDES $\rightarrow$ RSTPReid} \\
    \cmidrule(lr){4-7} \cmidrule(lr){8-11}
    \multirow{-2}{*}{Setting} & \multirow{-2}{*}{Method} & \multirow{-2}{*}{Ref.} & R@1 & R@5 & R@10 & mAP & R@1 & R@5 & R@10 & mAP \\
    \hline\hline
    \multirow{7}{*}{Fully Supervised}
    & IRRA~\citep{cvpr23crossmodal} & CVPR'23
    & 73.38 & 89.93 & 93.71 & 66.13
    & 60.20 & 81.30 & 88.20 & 47.17 \\
    & RaSa~\citep{bai2023rasa} & IJCAI'23
    & 76.51 & 90.29 & 94.25 & 69.38
    & 66.90 & 86.50 & 91.35 & 52.31  \\
    & TBPS-CLIP~\citep{cao2024empirical} & AAAI'24
    & 73.54 & 88.19 & 92.35 & 65.38
    & 61.60 & 81.20 & 87.70 & 47.71 \\
    & RDE~\citep{qin2024noisy} & CVPR'24
    & 75.94 & 90.14 & 94.12 & 67.56
    & 65.35 & 83.95 & 89.90 & 50.88 \\
    & PTMI~\citep{lu2025prompt} & TIFS'25
    & 76.02 & 89.93 & 94.14 & 70.85
    & 63.35 & 81.95 & 89.10 & 52.46 \\
    & GAHR~\citep{qi2025granularity} & TIFS'25
    & 76.64 & 90.46 & 94.10 & 66.81
    & 68.85 & 86.50 & 91.10 & 53.60 \\
    & ICL~\citep{qin2025human} & CVPR'25
    & 78.18 & 91.63 & 94.83 & 69.58
    & 70.00 & 86.60 & 91.70 & 54.16 \\
    \hline
    \multirow{3}{*}{\makecell[c]{UDA-TBPR}}
    & POUF~\citep{Tanwisuth2023POUFPU}   & ICML'23
    & 20.50 & 39.39 & 49.03 & 18.48 & 36.60 & 61.80 & 73.15 & 28.28 \\
    & ReCLIP~\citep{Hu2023ReCLIPRC} & WACV'24
    & 33.48 & 55.25 & 64.26 & 29.19 & 50.35 & 73.25 & 81.20 & 42.75 \\
    & GCKD~\citep{Luo2025GraphBasedCK}   & AAAI'25
    & 52.70 & 72.76 & 80.15 & 45.97 & 59.95 & 79.05 & 85.95 & 49.68 \\
    \hline
    \rowcolor[HTML]{D7F6FF}
    {\raisebox{-0.5\height}{\shortstack[c]{\strut Human-Annotation-Free \\ \strut UDA-TBPR}}}
    & RaSa + STaG & Ours
    & 57.44 & 76.13 & 83.25 & 50.97
    & 55.70 & 76.95 & 84.15 & 46.88 \\
    \hline\thickhline

\end{tabular}
}}
\end{table*}

\begin{table*}[t]
\centering
\caption{
Comparisons with state-of-the-art methods using \textbf{RSTPReid} as the source domain and \textbf{CUHK-PEDES / ICFG-PEDES} as target domains.
We follow the same target-domain supervision settings and cross-setting reference protocol defined in Table~\ref{tab:c2all}.
Within the human-annotation-free setting, ``Backbone+STaG'' evaluates the compatibility of STaG with different retrieval architectures, and $\Delta$ denotes the absolute performance change relative to the corresponding backbone baseline.
* indicates results reproduced by us due to the unavailability of official source code.
}
\label{tab:r2all}
\providecommand{\db}[1]{\textbf{+#1}}
\small{
\resizebox{\linewidth}{!}{
\setlength\tabcolsep{6pt}
\renewcommand\arraystretch{1.2}
    \begin{tabular}{cllcccccccc}
    \hline\thickhline
     &  &  & \multicolumn{4}{c}{RSTPReid$\rightarrow$CUHK-PEDES} & \multicolumn{4}{c}{RSTPReid$\rightarrow$ICFG-PEDES} \\
    \cmidrule(lr){4-7} \cmidrule(lr){8-11}
    \multirow{-2}{*}{Setting} & \multirow{-2}{*}{Method} & \multirow{-2}{*}{Ref.}
    & R@1 & R@5 & R@10 & mAP & R@1 & R@5 & R@10 & mAP \\
    \hline\hline
    \multirow{7}{*}{Fully Supervised}
    & IRRA~\citep{cvpr23crossmodal} & CVPR'23
    & 73.38 & 89.93 & 93.71 & 66.13
    & 63.46 & 80.25 & 85.82 & 38.06 \\
    & RaSa~\citep{bai2023rasa} & IJCAI'23
    & 76.51 & 90.29 & 94.25 & 69.38
    & 65.28 & 80.40 & 85.12 & 41.29  \\
    & TBPS-CLIP~\citep{cao2024empirical} & AAAI'24
    & 73.54 & 88.19 & 92.35 & 65.38
    & 65.05 & 80.34 & 85.47 & 39.83 \\
    & RDE~\citep{qin2024noisy} & CVPR'24
    & 75.94 & 90.14 & 94.12 & 67.56
    & 67.68 & 82.47 & 87.36 & 40.06 \\
    & PTMI~\citep{lu2025prompt} & TIFS'25
    & 76.02 & 89.93 & 94.14 & 70.85
    & 66.54 & 81.50 & 86.50 & 46.05 \\
    & GAHR~\citep{qi2025granularity} & TIFS'25
    & 76.64 & 90.46 & 94.10 & 66.81
    & 68.69 & 82.83 & 87.40 & 42.10 \\
    & ICL~\citep{qin2025human} & CVPR'25
    & 78.18 & 91.63 & 94.83 & 69.58
    & 69.22 & 83.49 & 88.06 & 42.34 \\
    \hline
    \multirow{3}{*}{\makecell[c]{UDA-TBPR}}
    & POUF~\citep{Tanwisuth2023POUFPU}   & ICML'23
    & 17.84 & 35.62 & 46.72 & 16.30 & 21.27 & 37.87 & 46.66 & 11.04 \\
    & ReCLIP~\citep{Hu2023ReCLIPRC}      & WACV'24
    & 21.04 & 40.29 & 50.28 & 18.92 & 27.18 & 42.51 & 50.28 & 17.83 \\
    & GCKD~\citep{Luo2025GraphBasedCK}   & AAAI'25
    & 51.01 & 70.29 & 77.29 & 43.71 & 46.40 & 61.17 & 66.97 & 26.06 \\
    \hline
    \multirow{14}{*}{\makecell[c]{Human-Annotation-Free \\ UDA-TBPR}}
    & IVT~\citep{shu2023see} & ECCVW’22
    & 16.73 & 33.65 & 43.01 & 14.42 & 19.43 & 35.22 & 43.83 & 13.52 \\
    & POUF~\citep{Tanwisuth2023POUFPU}  & ICML’23
    & 16.41 & 33.26 & 43.29 & 15.10 & 18.33 & 34.80 & 44.09 & 11.31 \\
    & CFine~\citep{yan2022clip} & TIP’23
    & 20.79 & 38.76 & 48.93 & 15.77 & 24.83 & 39.49 & 46.92 & 17.04 \\
    & ReCLIP~\citep{Hu2023ReCLIPRC} & WACV'24
    & 20.36 & 38.57 & 48.63 & 20.15 & 26.32 & 39.93 & 47.78 & 16.30 \\
    & GCKD*~\citep{Luo2025GraphBasedCK} & AAAI'25
    & 39.15 & 58.76 & 67.30 & 34.92 & 42.23 & 58.29 & 65.24 & 25.03 \\
    \cline{2-11}

    & RaSa~\citep{bai2023rasa} & IJCAI'23
    & 42.85 & 61.97 & 69.79 & 35.64
    & 41.30 & 56.18 & 62.36 & 22.39 \\
    &\textbf{+ STaG} & Ours
    & 54.97 & 74.43 & 81.69 & 48.00
    & 45.10 & 62.16 & 69.58 & 22.73 \\
    \rowcolor[HTML]{D7F6FF}
    \cellcolor{white}
    & \textbf{Improvement $\Delta$} &
    & \db{12.12} & \db{12.46} & \db{11.90} & \db{12.36}
    & \db{3.80}  & \db{5.98}  & \db{7.22}  & \db{0.34} \\
    \cline{2-11}

    & IRRA~\citep{cvpr23crossmodal} & CVPR'23
    & 32.65 & 55.20 & 65.38 & 30.17
    & 32.35 & 49.68 & 57.74 & 20.57 \\
    & \textbf{+ STaG} & Ours
    & 46.44 & 67.54 & 75.88 & 41.01
    & 40.64 & 59.23 & 67.15 & 20.92 \\
    \rowcolor[HTML]{D7F6FF}
    \cellcolor{white}
    & \textbf{Improvement $\Delta$} &
    & \db{13.79} & \db{12.34} & \db{10.50} & \db{10.84}
    & \db{8.29}  & \db{9.55}  & \db{9.41}  & \db{0.35} \\
    \cline{2-11}

    & TBPS-CLIP~\citep{cao2024empirical} & AAAI'24
    & 32.09 & 53.57 & 63.28 & 29.40
    & 37.04 & 52.65 & 60.13 & 23.23 \\
    & \textbf{+ STaG} & Ours
    & 40.14 & 59.58 & 68.03 & 35.27
    & 44.14 & 61.17 & 68.13 & 24.19 \\
    \rowcolor[HTML]{D7F6FF}
    \cellcolor{white}
    & \textbf{Improvement $\Delta$} &
    & \db{8.05}  & \db{6.01}  & \db{4.75}  & \db{5.87}
    & \db{7.10}  & \db{8.52}  & \db{8.00}  & \db{0.96} \\
    \hline\thickhline

\end{tabular}
}}
\end{table*}

\begin{table}[t]
\centering
\caption{
\textbf{Cross-dataset generalization of CAE.}
The STaG variants adapted on ICFG-PEDES $\rightarrow$ CUHK-PEDES
are directly evaluated on RSTPReid without further adaptation.
Baseline denotes the source-only model trained on ICFG-PEDES.
}
\label{tab:cae_cross_dataset}
\setlength{\tabcolsep}{9pt}
\renewcommand{\arraystretch}{1.1}
\begin{tabular}{lcccc}
\hline\thickhline
Model & R@1 & R@5 & R@10 & mAP \\
\midrule
Baseline
& 44.50 & 67.00 & 76.85 & 33.45 \\
STaG w/o $\mathcal{L}_{\mathrm{CAE}}$
& 50.15 & 72.85 & 80.75 & 37.86 \\
STaG w/ $\mathcal{L}_{\mathrm{CAE}}$
& \textbf{50.55} & \textbf{73.50} & \textbf{81.45} & \textbf{37.92} \\
\hline\thickhline
\end{tabular}
\end{table}

\subsection{Comparative Analysis with Human-Annotation-Accessible Baselines}
\label{app:bridging_gap}

We include fully supervised and text-accessible UDA-TBPR results
in Tables~\ref{tab:c2all} to \ref{tab:r2all} as references across different target-domain
supervision settings. Fully supervised configurations use paired
target-domain image--text annotations, whereas the text-accessible
UDA-TBPR configurations use human-written target-domain descriptions
during adaptation. In contrast, STaG uses labeled source-domain data
and unlabeled target-domain images, with target-domain textual
supervision provided by MLLM-generated pseudo-descriptions rather
than human-written annotations. STaG therefore operates under a more
restrictive target-domain annotation-access requirement. Comparisons
with these reference groups contextualize retrieval performance
across supervision settings, rather than establish methodological
superiority under identical supervision.

Despite this restriction, STaG remains competitive in several
evaluated transfer settings. For example, on CUHK-PEDES
$\rightarrow$ RSTPReid, RaSa+STaG achieves 60.75\% Rank@1, compared
with 57.44\% for GCKD and 52.15\% for ReCLIP in the text-accessible
UDA-TBPR group of Table~\ref{tab:c2all}. These cross-setting observations
complement the component ablations in Sec.~\ref{sec:ablation} and the comparison
with plain pseudo-text adaptation in Appendix~\ref{app:plain_pseudotext}, which assess
the contributions of STaG within the human-annotation-free setting.

\subsection{Generalization Across Diverse Source Domains}
\label{app:generalization}

We further evaluate STaG with different datasets serving as the source domain.
With CUHK-PEDES as the source, STaG reaches 50.45\% Rank@1 on ICFG-PEDES and 60.74\% on RSTPReid.
When ICFG-PEDES is used as the source, STaG obtains 57.44\% Rank@1 on CUHK-PEDES and 55.72\% on RSTPReid.
The consistent gains across these transfer directions indicate that the effectiveness of STaG is not limited to a single source--target configuration.
At the same time, the varying absolute performance across datasets reflects the different levels of cross-domain difficulty.

\subsection{Architectural Robustness}
\label{app:completing_matrix}
Table~\ref{tab:r2all} completes our comprehensive evaluation matrix by utilizing RSTPReid as the source domain, successfully adapting it to both CUHK-PEDES and ICFG-PEDES. Since RSTPReid is significantly smaller in scale and possesses distinct imaging characteristics compared to the other two datasets, it imposes an extreme visual domain shift.

This table serves two crucial purposes.
\ding{182} it verifies the architectural agnosticism of STaG. By integrating our framework with three structurally distinct baselines (RaSa, IRRA, and TBPS-CLIP), we observe universal and consistent performance surges, proving STaG is a highly compatible ''plug-and-play'' solution.
\ding{183} it highlights STaG's remarkable rescue effect under severe domain degradation. When transferring from RSTPReid to CUHK-PEDES, the domain gap is so drastic that the strong baseline IRRA collapses to a mere 32.65\% R@1. However, with the integration of STaG, the performance experiences a massive rebound, yielding a remarkable absolute improvement of +13.79\% in R@1 and +10.84\% in mAP.
This demonstrates that when direct cross-domain inference suffers substantial performance degradation, STaG can effectively improve the representation space and retrieval performance.

\subsection{Cross-Dataset Generalization of CAE}
\label{app:cae_generalization}

To examine whether the distribution sharpening introduced by CAE causes over-specialization to the adaptation domain, we directly evaluate the ICFG-PEDES $\rightarrow$ CUHK-PEDES adapted model on RSTPReid without further adaptation.

As shown in Table~\ref{tab:cae_cross_dataset}, adding CAE consistently improves all retrieval metrics, increasing Rank@1 from $50.15$ to $50.55$.
This result provides evidence that CAE does not impair cross-dataset transferability in this setting, while slightly improving the generalization of the adapted representation.

\section{Visualization Analysis}
\label{app: visualization}

To complement the quantitative evaluations, we provide qualitative analyses from four perspectives: spatial cross-modal grounding, attention over generated pseudo-texts, neighborhood retrieval in HGMA, and final retrieval behavior.
These visualizations provide an intuitive view of how SATA and HGMA affect the learned representations under domain shift.

\subsection{Visualization of Image Activation Maps}
\label{app:vis_image}

\begin{figure}[t]
  \centering
  \setlength{\abovecaptionskip}{4pt}
  \includegraphics[width=0.9\linewidth]{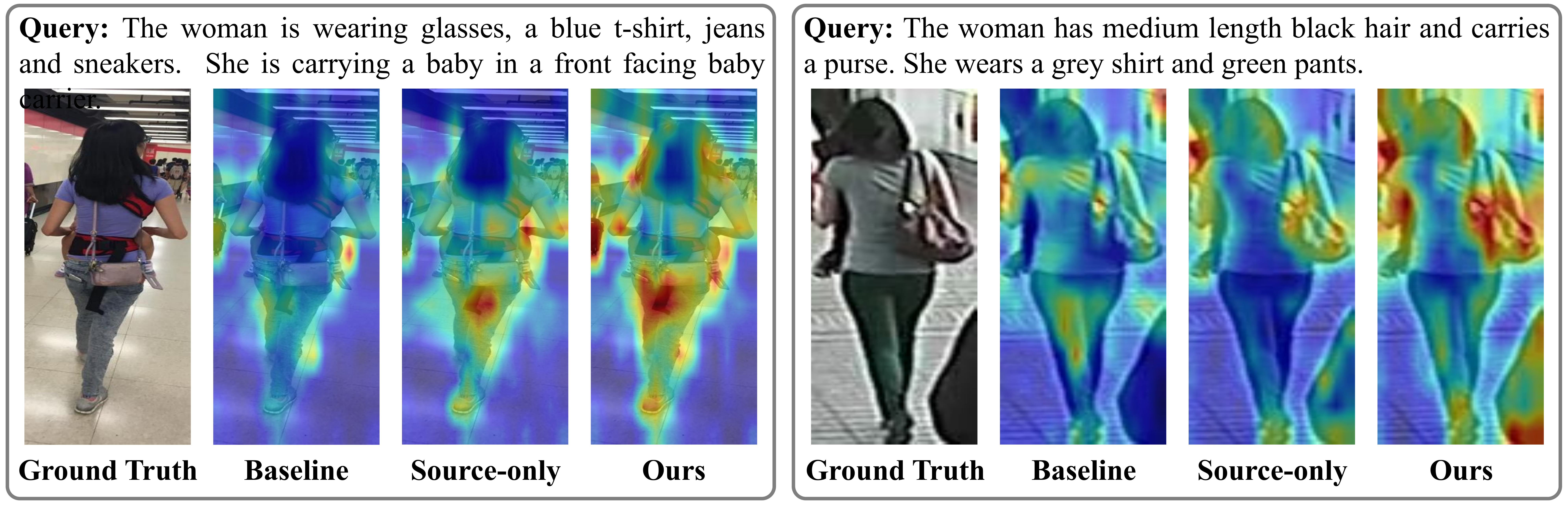}
    \caption{\textbf{Visualization of activation maps and token-level importance.}
    We compare three configurations on representative positive image–text pairs for the transfer ICFG-PEDES $\rightarrow$ CUHK-PEDES:
    \textbf{Baseline} denotes TBPS-CLIP~\citep{cao2024empirical} trained on the source domain and directly evaluated on the target domain without any adaptation;
    \textbf{Pseudo-Text Adaptation} denotes TBPS-CLIP directly fine-tuned with target-domain image--pseudo-text pairs without SATA or HGMA; \textbf{Ours} denotes the full STaG framework. The \textbf{Ground Truth} column shows the original target-domain image for reference. Warmer colors indicate stronger spatial activation.}
  \label{fig:att-half}
\end{figure}

\begin{figure}[t]
  \centering
  \setlength{\abovecaptionskip}{4pt}
  \includegraphics[width=0.9\linewidth]{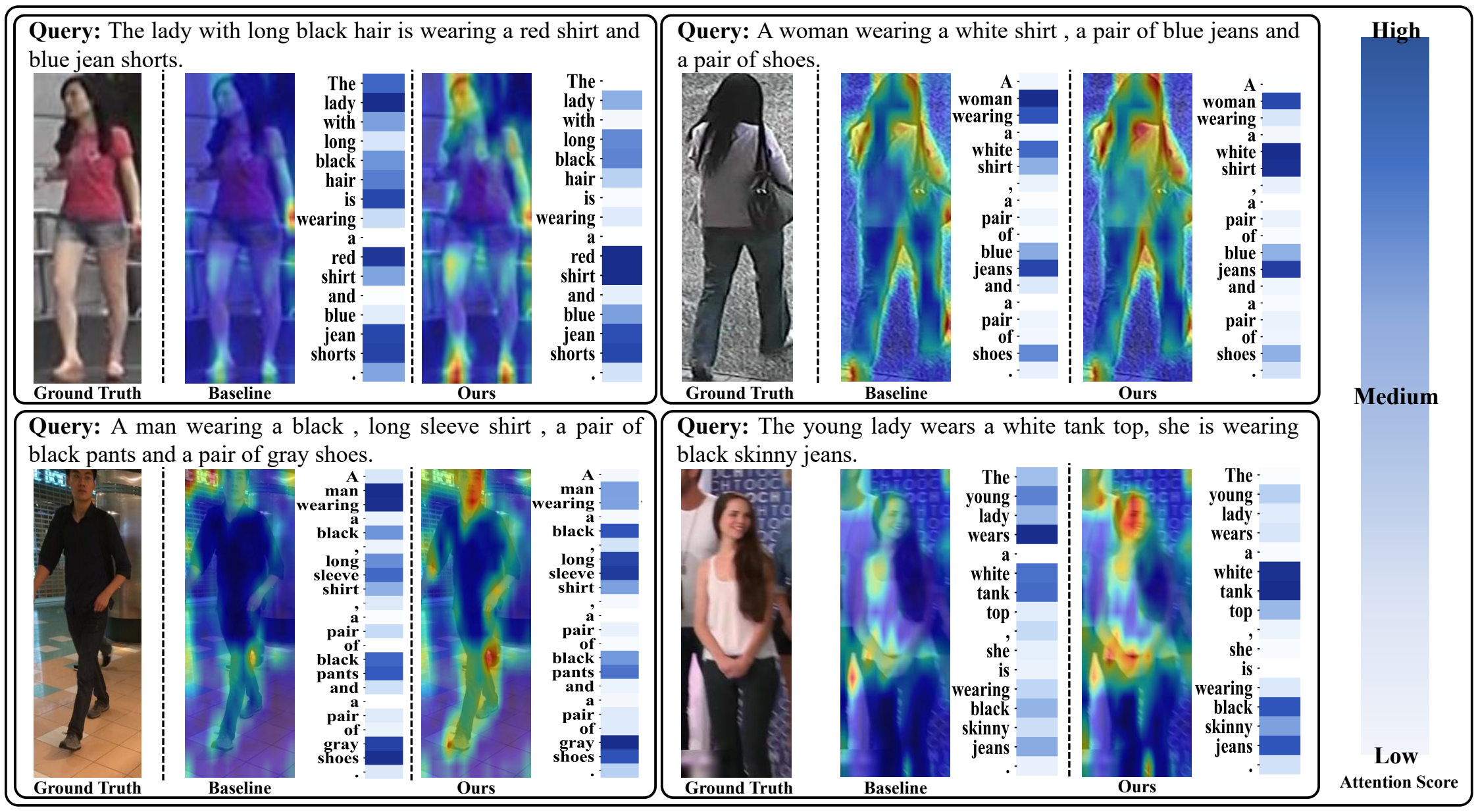}
  \caption{\textbf{Visualization of spatial activation and token-level importance.}
  We compare TBPS-CLIP~\citep{cao2024empirical} (Baseline) with STaG (Ours) on representative positive image-text pairs under the human-annotation-free UDA-TBPR setting, using ICFG-PEDES as the source domain and evaluating on target datasets. Warmer colors indicate stronger spatial activation on the image, and the bar plots show the normalized contribution of each word to the matching score.}
  \label{fig:app_img_atten}
\end{figure}

Figs.~\ref{fig:att-half} and~\ref{fig:app_img_atten} visualize the spatial and textual attention learned under domain shift.
Compared with the baseline, STaG consistently produces more concentrated responses on query-relevant person regions, such as upper-body clothing, pants, and shoes, while reducing attention to less relevant visual regions.
A similar tendency is observed on the text side, where attribute-bearing words related to colors, garments, and accessories contribute more strongly to the matching score.

The consistency between spatial localization and textual importance suggests that STaG improves not only global feature alignment, but also the correspondence between fine-grained visual evidence and descriptive attributes.
These observations provide qualitative support for the improved cross-modal grounding achieved by STaG under domain shift.

\begin{figure}[t]
  \centering
  \setlength{\abovecaptionskip}{4pt}

  \includegraphics[width=0.9\linewidth]{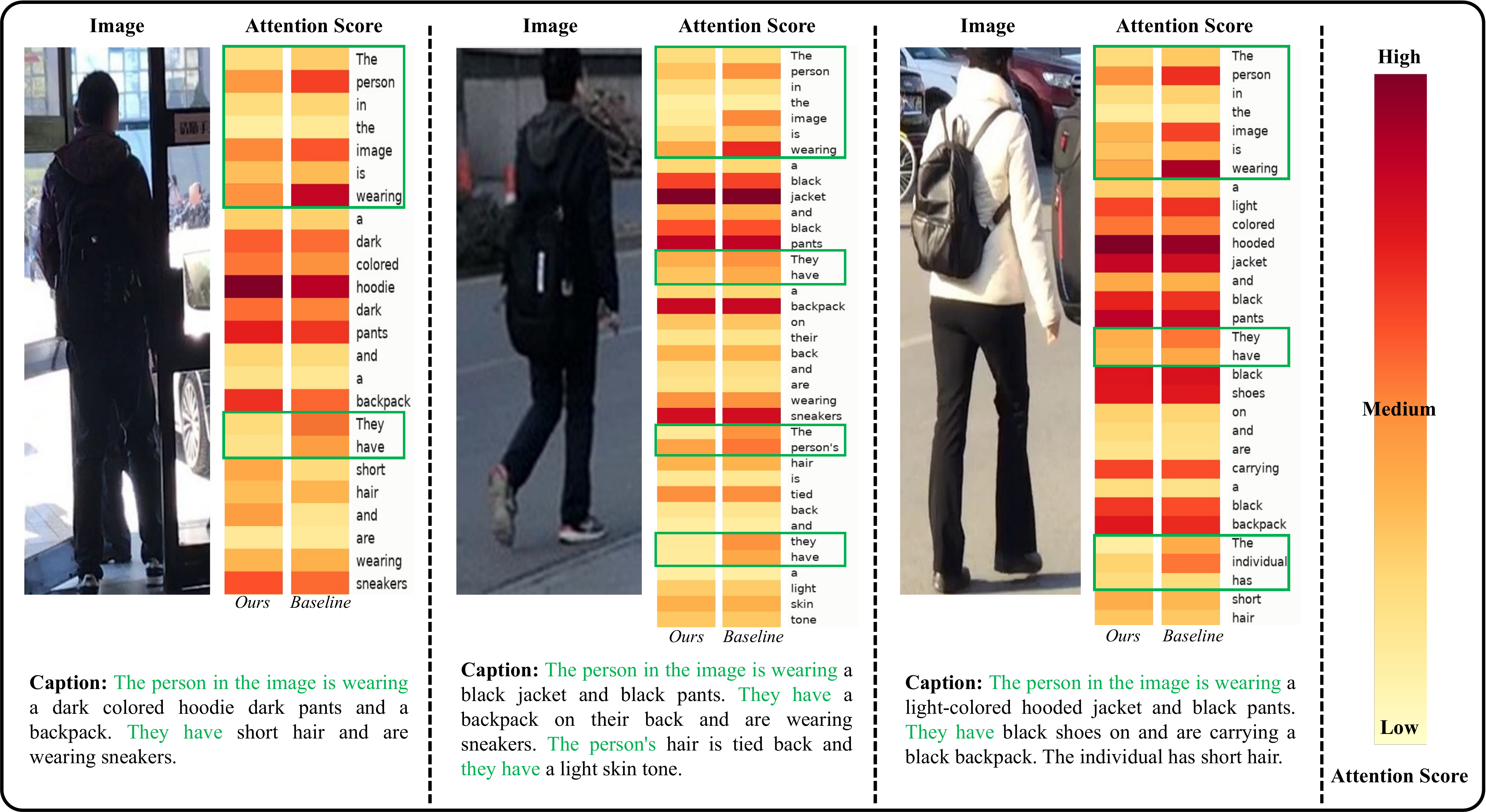}
  \caption{\textbf{Token-level attention on generated pseudo-texts.}
  We compare TBPS-CLIP~\citep{cao2024empirical} (Baseline) with STaG (Ours) on representative image--pseudo-text pairs under the human-annotation-free UDA-TBPR setting (ICFG-PEDES as source). Bar plots show the normalized contribution of each token to the matching score. Green boxes mark MLLM-specific stylistic templates, which carry no identity-discriminative information. The baseline distributes attention diffusely across these templates, whereas STaG concentrates attention on attribute-bearing tokens.}
  \label{fig:app_txt_atten}
\end{figure}

\subsection{Visualization of Attribute-Level Attention in Pseudo-Texts}
\label{app:att in text}

To complement the quantitative style-invariance analysis in Sec.~\ref{ablation:sata}, Fig.~\ref{fig:app_txt_atten} visualizes token-level attention over generated pseudo-texts.
The baseline assigns noticeable attention to recurring MLLM-specific templates such as ``The person in the image is wearing...'', whereas STaG shifts attention toward attribute-bearing words describing colors, garments, and accessories.

These template expressions occur frequently across generated descriptions but provide limited instance-specific visual information.
Reducing their influence therefore helps prevent the model from relying excessively on generator-specific writing patterns.
Meanwhile, the increased attention to observable attributes is consistent with the motivation of SATA: using shared visual anchors to encourage representations that preserve semantic content while reducing sensitivity to textual style.
This visualization provides instance-level evidence complementary to the distribution-level analysis in Appendix~\ref{app: distribution}.

\subsection{Visualization of Retrieved HGMA Neighbors}
\label{app:hgma_neighbors}

\begin{figure}[t]
  \centering
  \setlength{\abovecaptionskip}{4pt}
  \includegraphics[width=0.85\linewidth]{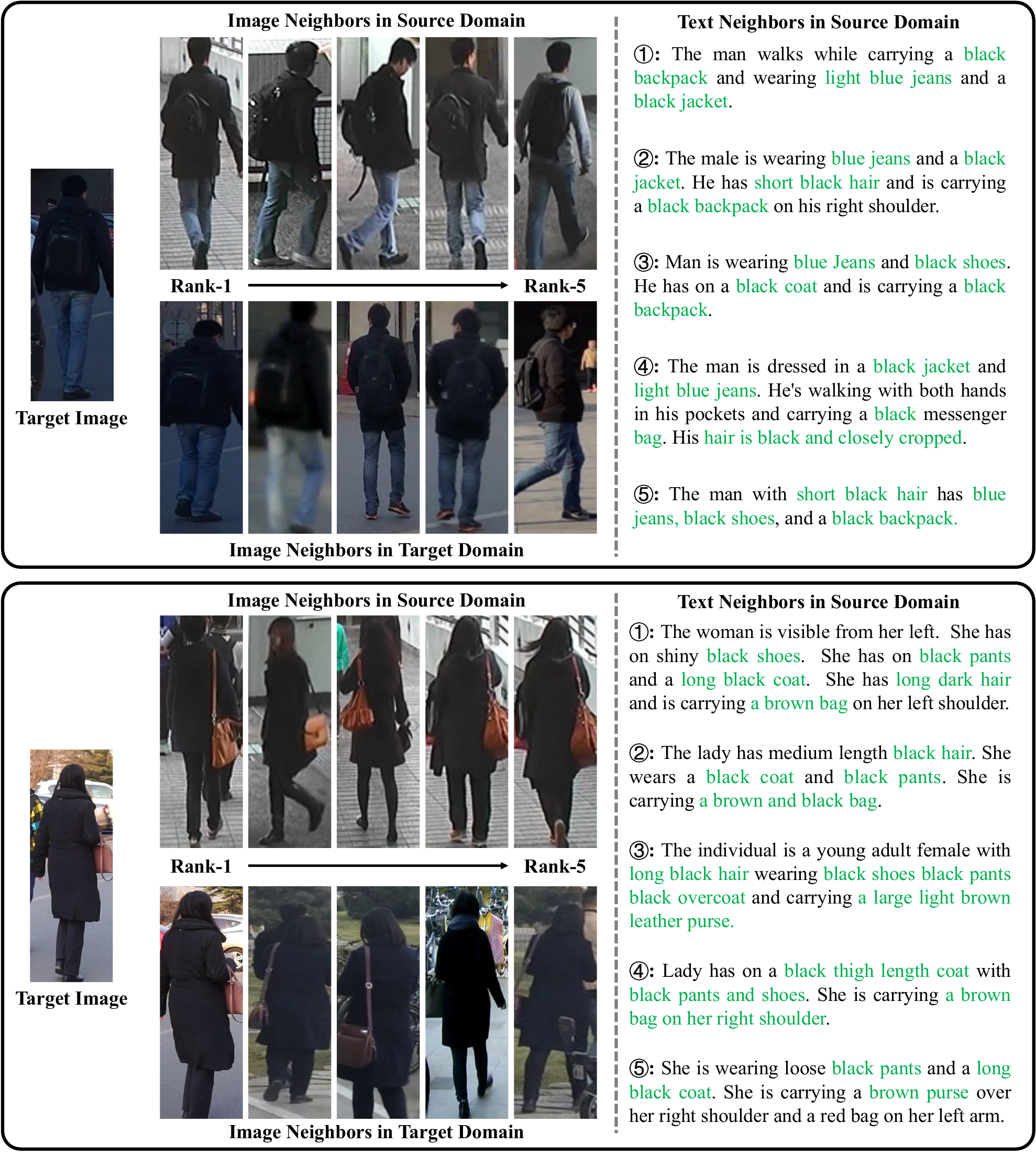}
    \caption{
    \textbf{Visualization of HGMA-retrieved neighbors.}
    Given a target query, HGMA retrieves feature-similar anchors from the source-image, source-text, and target-image memories.
    Target-domain annotations, if displayed, are used only for post-hoc visualization and are not accessed during adaptation.
    }
    \label{fig:hgma_neighbors}
\end{figure}

Fig.~\ref{fig:hgma_neighbors} visualizes representative neighbors retrieved from the source-image, source-text, and target-image memories.
Although the cross-domain samples do not share identities with the target query, the retrieved anchors exhibit similar fine-grained appearance attributes, such as clothing color, garment type, and carried objects.

The three memories provide complementary cues: source-image neighbors offer cross-domain visual references, source-text neighbors introduce text-side semantic information, and target-image neighbors preserve the local structure of the target domain.
Their semantic consistency illustrates how HGMA combines heterogeneous but feature-similar evidence rather than relying on assumed cross-domain identity correspondence.
Together with the random-neighbor and memory-source ablations, this visualization qualitatively supports the neighborhood construction adopted by HGMA.

\subsection{Visualization of Retrieval Results}
\label{app:vis_retrieval}
\begin{figure}[t]
  \centering
  \setlength{\abovecaptionskip}{4pt}

  \includegraphics[width=1\linewidth]{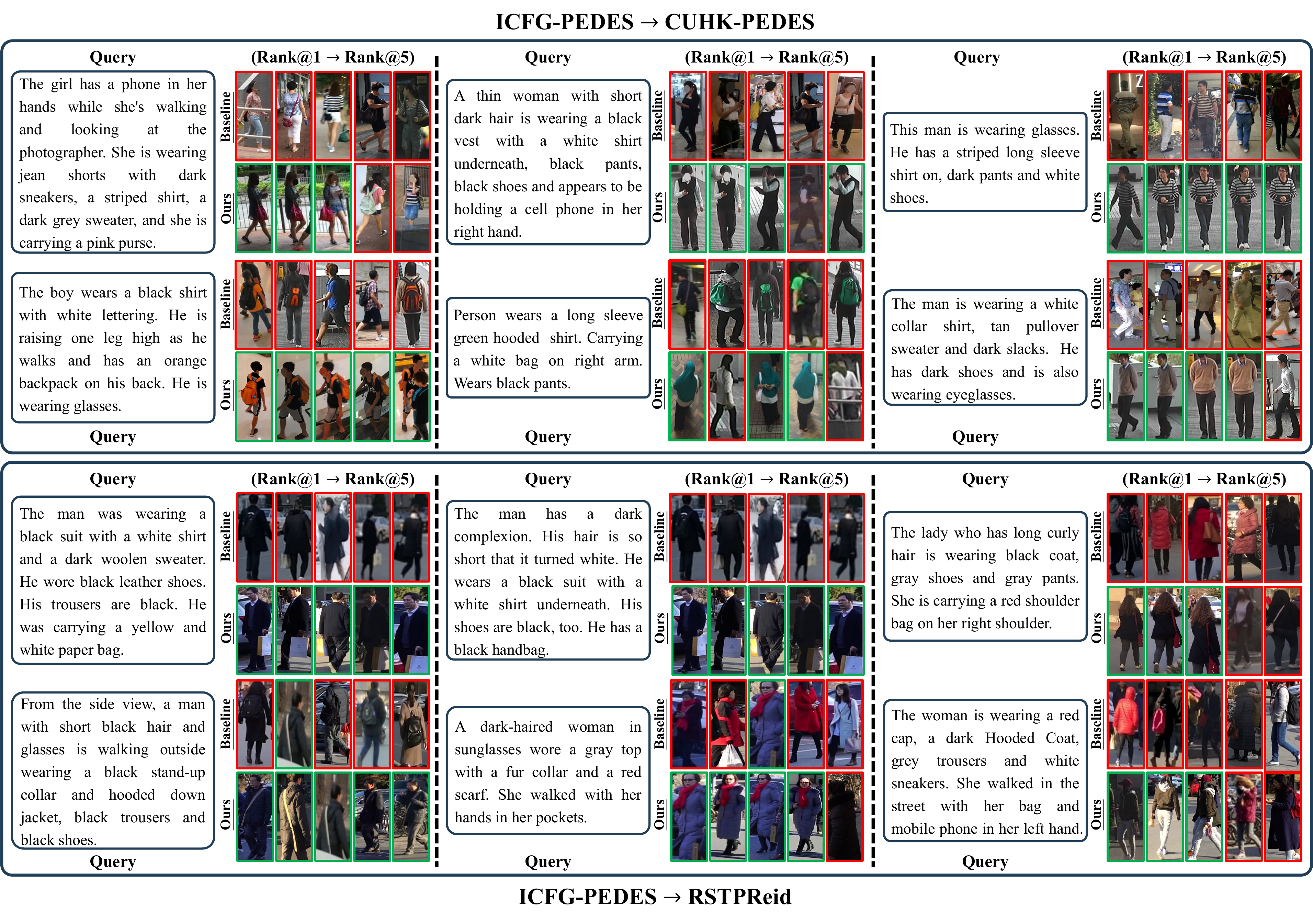}
  \caption{\textbf{Qualitative retrieval comparison (Rank@1$\rightarrow$Rank@5).}
  We show the top-5 retrieved images of TBPS-CLIP~\citep{cao2024empirical} (Baseline) and STaG (Ours) for representative text queries under the human-annotation-free UDA-TBPR setting.
  Green and red boxes denote correct and incorrect matches, respectively.}
  \label{fig:app_img_rank}
\end{figure}

Fig.~\ref{fig:app_img_rank} presents representative top-5 retrieval results under two cross-domain transfer settings.
Compared with the baseline, STaG places more correct matches at higher ranks and reduces errors caused by visually similar distractors.
The baseline often retrieves incorrect candidates that share a dominant coarse attribute with the query, such as clothing color, while differing in other fine-grained cues such as garment type, footwear, or accessories.

In contrast, STaG better preserves combinations of multiple descriptive attributes and remains more stable under variations in illumination, viewpoint, and background.
These examples suggest that the quantitative improvements in Rank@K and mAP are accompanied by more discriminative ranking behavior rather than merely changes in overall similarity scores.

\end{document}